# World of SWECraft: Agent Collaboration at Scale via Simulated Software Worlds

**Shannon Zejiang Shen**[1,*] **Yapei Chang**[2,*] **Qiyao Liang**[1,*] **Benjamin Newman**[3,*] **Jenny Liang**[4,*] **Rulin Shao**[3,*]
**Ziqiao Ma**[5] **Junjie Oscar Yin**[3] **Yung-Sung Chuang**[1] **Weijia Shi**[3]
**Yejin Choi**[6] **Luke Zettlemoyer**[3] **Pang Wei Koh**[3] **Mohit Iyyer**[2] **Tongshuang Wu**[4] **David Sontag**[1]

[1]Massachusetts Institute of Technology [2]University of Maryland, College Park [3]University of Washington
[4]Carnegie Mellon University [5]University of Michigan [6]Stanford University

`★ Core contributors` `zjshen@mit.edu` `szj.io/sw`

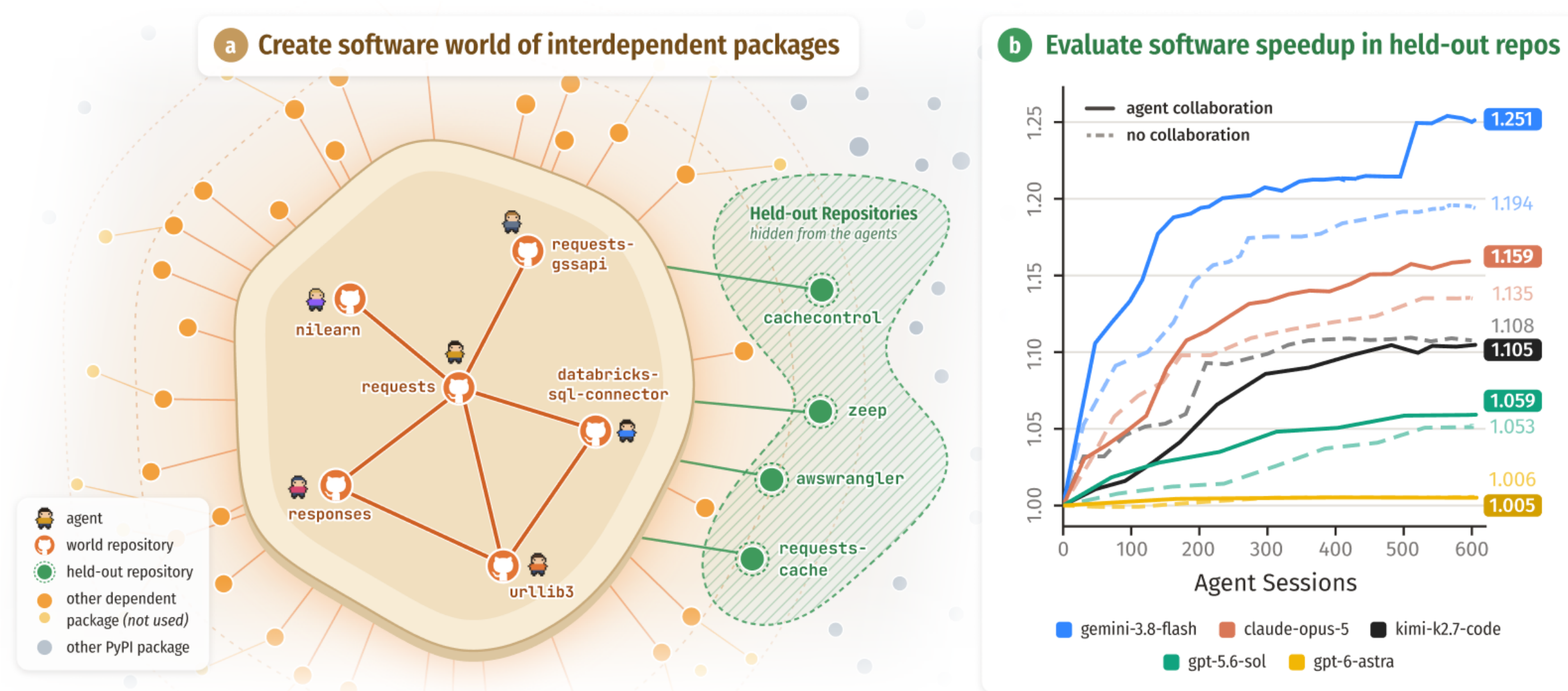

**Figure 1 Overview of Software World.** Left: we build a software world from real Python libraries that depend on one another. Agents maintain these libraries and work together to find tasks and make the code faster, and we measure the speedup on held-out repositories that depend on libraries in the world. Right: held-out speedup by agent sessions, with collaboration (solid) and without (dashed). `gemini-3.8-flash` and `claude-opus-5` agents perform better when collaboration is enabled, while the effect of collaboration is unclear for `kimi-k2.7-code` and `gpt-5.6-sol`. `gpt-6-astra` exhibits less proactivity and stops discovering tasks prematurely despite its strong general capabilities.

## Abstract

As teams of AI agents take on larger projects and long-standing open problems, agent collaboration offers another dimension for scaling inference-time compute. To understand what makes such collaboration effective and scalable, we develop Software World, a framework for simulating multi-agent coworking in realistic software ecosystems. It comes with a pipeline to create environments of real, interdependent Python packages of different topics, sizes, and dependency structures, which we call *software worlds*. In a simulation, agents maintain packages in the world with a high-level goal of improving software efficiency, and they are free to discover tasks to work on and collaborate with others. We evaluate agents' work in a world collectively by measuring speedup in hidden dependent packages. The framework allows us to study different elements of collaboration by varying communication, ownership, and resource allocation. We run simulations in three different worlds, totaling more than 10,000 hours of agent work across eight models. Overall, performance improvement scales with inference-time compute, with the best model improving downstream efficiency by up to 31%. We show that most agents can find meaningful tasks via collaboration, but some frontier models can prematurely stop looking for improvements and make little progress. We release the simulation framework and traces to support future studies of agent collaboration at `https://szj.io/sw`.

## 1 Introduction

Scaling agent teams has enabled AI systems to make substantial progress on problems that are difficult and time-consuming even for human experts. For example, recent efforts have used 16 agents to build a C compiler capable of compiling Linux (Carlini, 2026), dozens to formalize Fermat's Last Theorem (Anthropic, 2026), and approximately 10,000 working concurrently to develop a reported proof addressing the Navier–Stokes Millennium Prize problem (OpenAI, 2026). As these efforts expand from a few agents to thousands, understanding *how* agents work together through large-scale collaboration becomes increasingly important.

To study large-scale collaboration, we introduce Software World,[1] a framework for simulating agents maintaining real repositories and robustly evaluating their collective software improvements. We construct *software worlds* from interdependent packages and instantiate agents to maintain them together. Given a broad goal to improve software efficiency, agents discover and choose tasks themselves. They are free to work with others, communicating through issues and pull requests as human developers do. The world's collective outcome is evaluated with extrinsic signals—using tests and benchmarks from held-out dependent packages that are not visible to the agents during simulation.

**Software World leverages realistic setups for studying collaboration at scale.** Human developers already organize into communities and coordinate their work across interdependent projects (Cataldo et al., 2006; Bogart et al., 2016). Unlike completing an assigned issue (Jimenez et al., 2024) or solving predefined problems (Kim et al., 2025; Pappu et al., 2026; Park et al., 2026), collaboration between maintainers results in *discovering new tasks*, e.g., fixing bugs reported by others, or requesting new features from dependent libraries. It creates sufficient scope for agents to divide tasks and exchange information for meaningful collaboration. Compared to studies using games and virtual worlds to study collaboration qualitatively (Park et al., 2023), Software World also provides reliable quantitative measures that reflect the practical value of agents' collaboration outcomes.

**Software World provides configurable controls for understanding elements of collaboration.** Following Bratman (1992), we consider collaboration a form of shared *agency*: two or more agents intentionally organize and coordinate their actions as contributions to a joint activity. Collaboration involves *communication* among *task owners* to coordinate their work and *allocate resources* toward a shared goal (Malone and Crowston, 1994). Existing studies examine these aspects in isolation, e.g., focusing on communication protocols (Park et al., 2026; Pappu et al., 2026), or configuring agent tasks and roles through specialized harnesses and fixed workflows (Fourney et al., 2024; Hong et al., 2024; Pan et al., 2025). In a software world, we can systematically control communication access, code ownership, and resource allocation to study how these conditions shape collaboration and impact the resulting work.

**We simulate software worlds with 10,000 hours of agent work.** We create three representative software worlds and run simulations across eight models with 55K agent sessions totaling more than 180B tokens. Overall, we find that performance improvement scales with the compute an agent spends in its work, and collaboration helps agents discover what to work on, with the best model improving downstream efficiency by up to 31%. We see signs that break the "curse of coordination": capable agents can meaningfully collaborate when there is clear task ownership. Surprisingly, some frontier models can stop looking for improvements and make little progress, and agents struggle to effectively allocate shared resources within the world.

Our work contributes (1) Software World, a framework for simulating and evaluating collaboration among agents maintaining interdependent software packages; (2) empirical findings on how communication, ownership, and resource allocation shape agent behavior and collective performance; and (3) an open dataset of 40K agent session traces from the main experiments.

[1]Our title, "World of SWECraft," plays on *World of Warcraft*, where SWE stands for software engineering. *World of Warcraft* is a massively multiplayer online role-playing game (MMORPG), where players explore a shared fantasy world and cooperate on quests and battles. Similarly, in our software worlds, multiple agents are launched, free to discover tasks and collaborate with each other.

## 2 The Software World Framework

A software world contains interdependent packages, agents that maintain them, and records of their communication. Software World provides two components for studying collaboration in these worlds. SW-Env constructs the worlds and runs agents under configurable working conditions (Section 2.1). SW-Eval measures the effects of their changes on separate dependent packages (Section 2.2).

### 2.1 SW-Env: The Software World Environment

We give agents a broad goal of improving software efficiency, then let them choose what to work on and when to involve other maintainers. We construct software worlds with interdependent packages to create opportunities for collaboration, since a problem one agent discovers may require another agent to fix it. Their changes and discussions persist, so agents can build on earlier work and respond as other maintainers change the software. Below, we describe how we construct and run these worlds.

**Constructing software worlds.** We develop a pipeline for selecting repositories from real Python packages. It identifies *core packages* with foundational roles in the ecosystem, such as `pandas` or `requests`, then selects *consumer packages* that use them in different ways and form different dependency structures. We manually review the resulting selections to form software worlds (see details in Appendix A). Section 3 presents worlds with different domains, sizes, and dependency structures. This pipeline could also be applied to other programming languages, although our experiments focus on Python.

**Agents and work sessions.** In our setup, an agent is an LLM with tools and persistent memory. It is given a broad goal of improving software efficiency while preserving correctness, and we let it choose what to work on. Agents can inspect and edit code, run tests and profilers, exchange messages, and publish releases. They can also decide that no useful work remains. We extend the Harbor framework (Harbor Framework Team, 2026) to run agents over multiple *sessions*. Each session gives an agent a fresh sandbox with access to a Bash shell and lasts up to a configured time limit. The instructions ask agents to review their saved memory and message notifications before planning their work, then summarize what they did and update their memory and work log at the end. Code, discussions, and saved memory carry over between sessions, while the conversation starts fresh. Empirically, we find this design outperforms launching one long session per agent, as it avoids an indefinitely growing conversation and simplifies context management (Appendix C). Each agent runs one session at a time, and multiple agents can work in parallel up to a concurrency limit.

**A shared platform for code and discussion.** We coordinate agent work through Forgejo, a software development platform similar to GitHub, where each package has its own repository. We import each package's source code, commit history, and tags, but leave out existing issues and pull requests. All subsequent discussions and change proposals come from the agents. Agents discuss problems through issues and propose and review code changes through pull requests. They can continue working after sending a message without waiting for a reply. Each agent receives a unique account through which the platform enforces read/write permissions for the repositories.

Within each world, we can change which repositories agents can inspect and modify, who they can communicate with, and how they receive and share work-session budgets. Our later experiments vary these settings to study *communication*, *ownership*, and *resource allocation* (Sections 4 to 6).

### 2.2 SW-Eval: Evaluating Software Worlds

Agents' improvements to a library in the world can benefit other software that depends on it. We use this dependency relationship to evaluate the downstream impact of agents' work. SW-Eval evaluates whether agents' changes make dependent software faster while preserving correctness, using human-authored benchmarks that agents never see during their work sessions.

**Extrinsic evaluation.** We select held-out packages to *extrinsically* evaluate whether agents' changes improve software beyond the repositories inside a world. These packages depend on the core libraries in the world, but agents are not told which packages and workloads are selected for evaluation and cannot modify the evaluator's copies. Held-out packages come with human-authored software efficiency benchmarks whose workloads simulate their realistic usage.[2] Unlike workloads that directly benchmark the core libraries, these workloads measure how upstream improvements propagate through the dependency chain to benefit downstream software.

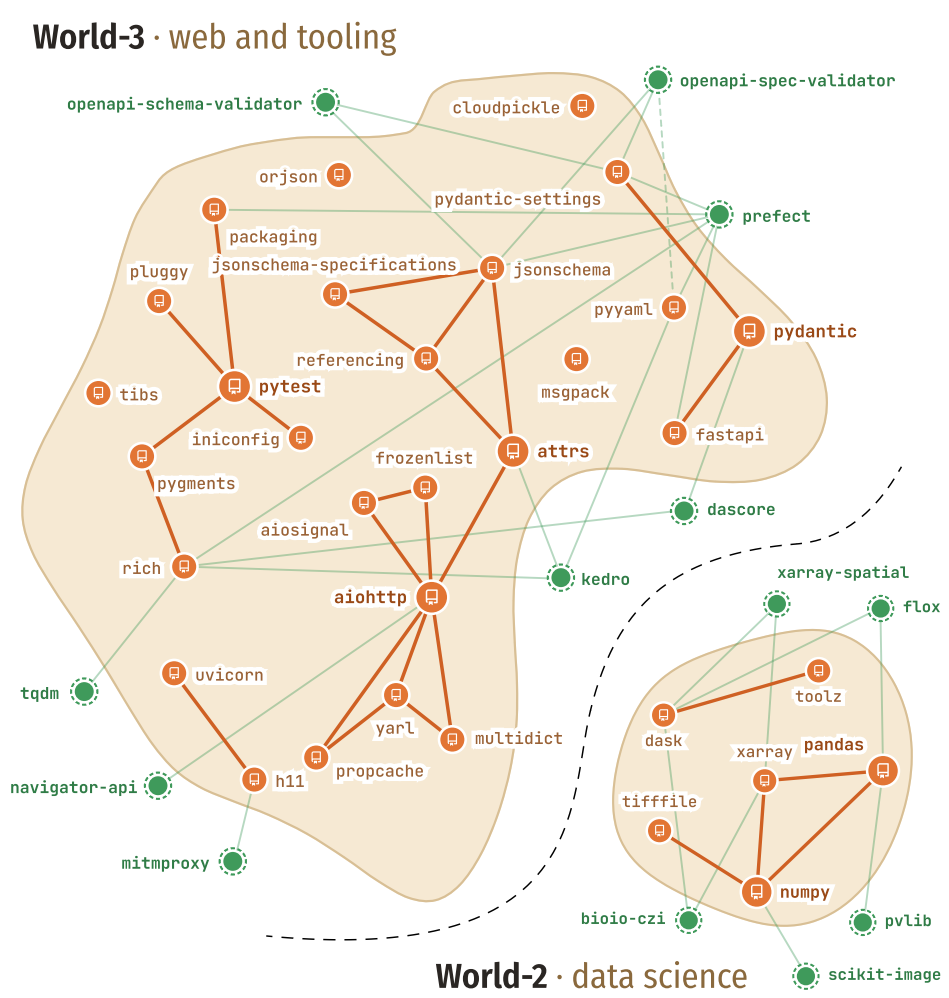

| | World-1 | World-2 | World-3 |
|---|---|---|---|
| Domain | networking | data science | web and tooling |
| Core libraries | requests, urllib3 | numpy, pandas | pydantic, attrs, aiohttp, pytest |
| Packages | 6 | 6 | 26 |
| Lines of code (w/o tests) | 165k | 601k | 329k |
| Technical stack | Pure Python | C + Cython | Rust + C |
| Monthly downloads | 2.5 B | 1.4 B | 13.7 B |
| Dependency edges | 7 | 5 | 22 |
| Dependency clusters | 1 | 2 | 9 |
| Society structure | Star-like | Star-like | Multi-center |
| Held-out packages | 4 | 5 | 8 |
| Benchmark tasks | 17 | 17 | 20 |
| In-world packages evaluated | 2 of 6 | 5 of 6 | 13 of 26 |

**Figure 2 The three software worlds.** Left: illustrations of packages inside World-2 and World-3 and their dependencies (orange), with the held-out packages that depend on them (green). Packages in World-1 are illustrated in Fig. 1. Right: detailed information on the structure, libraries, and statistics of each world.

**Measuring software speedups.** We measure software speedups via simulated CPU instruction counts with `pytest-codspeed` and Callgrind (Shypula et al., 2024).[3] We run evaluations at regular checkpoints of the world's state: per 50 sessions, we snapshot the latest state of repositories in the world, and run the benchmarks after installing the exact version of core libraries in fixed environments for the held-out packages. Let $I_t(k)$ denote the instruction count for task $t$ at the $k$-th evaluation checkpoint; its measured speedup is $r_t(k) = I_t(0)/I_t(k)$, with $I_t(0)$ being the initial count. We aggregate speedups equally across all held-out tasks $\mathcal{T}$ using the geometric mean (Fleming and Wallace, 1986):

$$S(k) = \sqrt[|\mathcal{T}|]{\prod_{t \in \mathcal{T}} r_t(k)}.$$

**Accounting for correctness failures.** However, as agents may introduce changes that break downstream software, purely measuring performance changes may fail to capture this issue. We therefore use each held-out package's correctness checks to determine how its tasks enter the aggregate score. Let $\tilde{r}_t(k)$ denote the credited speedup, and $\mathcal{T}_H$ denote the tasks from the held-out package $H$. If package $H$ passes its checks at checkpoint $k$, each task $t \in \mathcal{T}_H$ receives its measured value, $\tilde{r}_t(k) = r_t(k)$. If its checks fail, we consider two policies:

1. *Strict:* assign $\tilde{r}_t(k) = 1$ for every $t \in \mathcal{T}_H$, representing no credited improvement.
2. *Carry:* retain $\tilde{r}_t(k) = r_t(k')$, where $k' < k$ is the most recent checkpoint at which $H$ passed its checks.

For each policy, we compute the aggregate score by replacing $r_t(k)$ with the credited speedup $\tilde{r}_t(k)$ in the geometric mean above, yielding $S_{\text{strict}}$ and $S_{\text{carry}}$. $S_{\text{strict}}$ makes the introduction and repair of correctness failures visible in the score, while $S_{\text{carry}}$ is closer to realistic software adoption, which retains the latest valid upstream release. We use carry for the main results.

[2] We trace execution to ensure that selected workloads actually require in-world code, since benchmarks can declare dependencies without actually using them (Appendix A).

[3] We choose this over wall-clock time, since wall-clock time can vary substantially across machines and execution conditions (Chen et al., 2026d). Appendix C reports how stable both measures are in our evaluations.

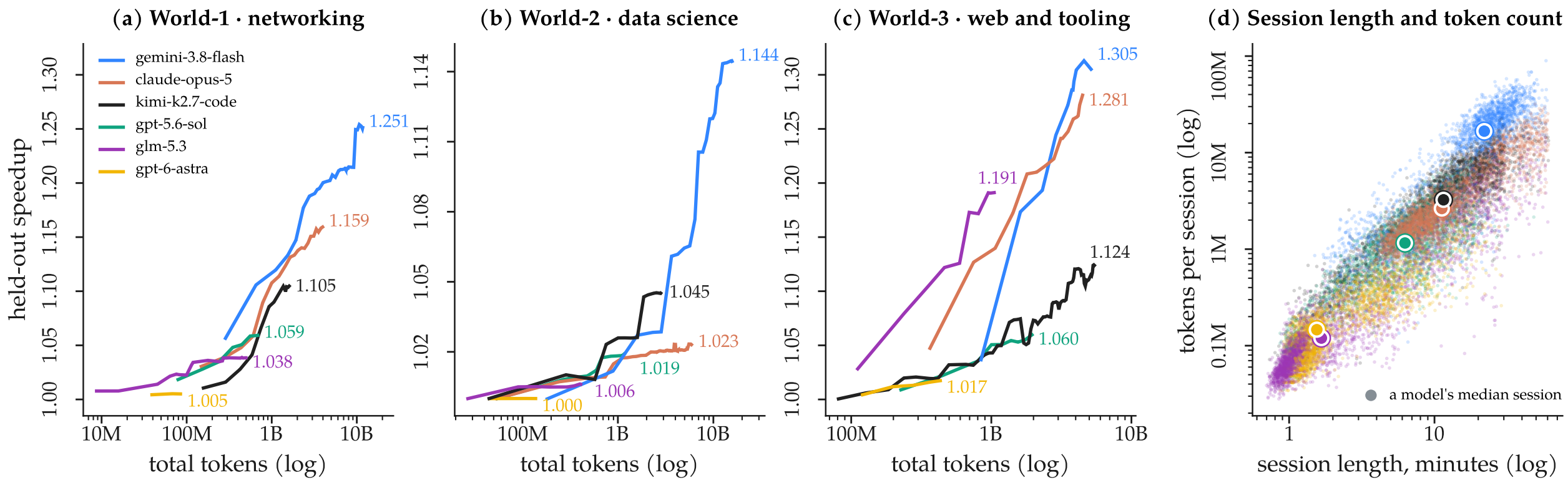

**Figure 3 Held-out speedup across three software worlds.** (a)–(c) Aggregate downstream speedup versus total tokens used by six models. Overall, efficiency improvements scale with the number of tokens agents use, with substantial differences across models; `gemini-3.8-flash` achieves the highest final score in each world. (d) Duration and token usage of agent sessions, with each model's median session highlighted. Session lengths vary across models, with `gemini-3.8-flash` having the longest median duration.

# 3 Simulating Three Software Worlds

We use Software World to construct three software worlds with different domains, sizes, and dependency structures, and we run agents from several model families in each. We first describe the worlds and the simulation setup, and then examine the resulting software improvements and the agents' work patterns.

## 3.1 Experimental Setup

**Software worlds.** We create three worlds to study how agents collaborate across different types of software and dependency structures (Fig. 2). World-1 contains six networking packages built around `requests` and `urllib3`, where agents optimize pure Python code. World-2 contains six data-science packages, including `numpy` and `pandas`, where agents can work with compiled code in addition to Python. World-3 expands to 26 packages across several dependency clusters, testing agents in a larger, more varied world.

**Models and harnesses.** We run simulations with both open and closed models from five families, namely, Claude, GPT, Gemini, Kimi, and GLM. We use the Pi coding agent harness whenever possible, as we find that it reduces cost while maintaining performance (Appendix C).[4] We use the default medium reasoning level of the harness for each model, and disable subagents within sessions so that agents cannot create additional workers outside the experiment's communication and ownership settings. For all runs, we record the token counts of each turn and estimate API costs based on the officially listed prices.

**Simulation setup.** By default, each agent is assigned to maintain one repository; agent communication is allowed, and agents are encouraged (though not forced) to reach out to and discuss with others. We run each agent initially for 100 sessions and let it continue if it is still working. We stop when it decides there is nothing left to do, or when the simulation reaches 10B tokens. We track agents' repository activity through pull requests, releases, issues, and comments. Sessions without these actions may still include local edits, tests, or updates to notes.

## 3.2 Simulation Results

**Performance improvement scales with compute.** Overall, downstream efficiency keeps improving with the compute spent. Given an open-ended goal and no prescribed tasks, agents from different models act autonomously: they make plans and change the code over hundreds of sessions. The agents' code changes generalize to the held-out packages (Fig. 3), with substantial improvements in downstream efficiency: the best run in each world reaches a speedup between 1.14 and 1.31. Across worlds, we see smaller improvements in World-2, whose core libraries are more complex and mix Python with C and C++.

[4]We run Opus on Claude Code, because Anthropic limits Claude Code subscriptions to its official harness.

**Frontier models can stop looking for work despite room for improvement.** `gemini-3.8-flash` achieves the highest held-out score in all three worlds, far ahead of recent frontier models like `gpt-6-astra` and `claude-fable-5-1` (Fig. 3). It also has the longest median session duration of 21.8 minutes and is the most proactive in seeking work to do, as shown in Fig. 3(d). By contrast, several frontier models stop working early because they conclude that nothing remains to be done. For example, `claude-opus-5-5` and `claude-fable-5-1` idle for more than 90% of their allotted sessions (Fig. 4). `gpt-6-astra` has no recorded repository activity in 91% of its completed sessions, including its final 100 sessions.

We further test these models with a *discovery prompt* that explicitly encourages them to explore before concluding that no work remains (Appendix B.2). Figure 4 shows that the models respond differently. `claude-opus-5-5` and `claude-fable-5-1` roughly double their work sessions (207 to 409 and 202 to 417), with held-out scores rising from 1.055 to 1.092 and 1.021 to 1.060, respectively.[5] Both models improve early, then plateau, spending about 86% of completed sessions idle. For `gpt-6-astra`, sessions without recorded repository activity fall from 91% to 56%, but its score remains nearly unchanged (1.005 to 1.003). These results suggest that frontier models with strong capabilities can struggle when they must decide for themselves what is worth doing next.

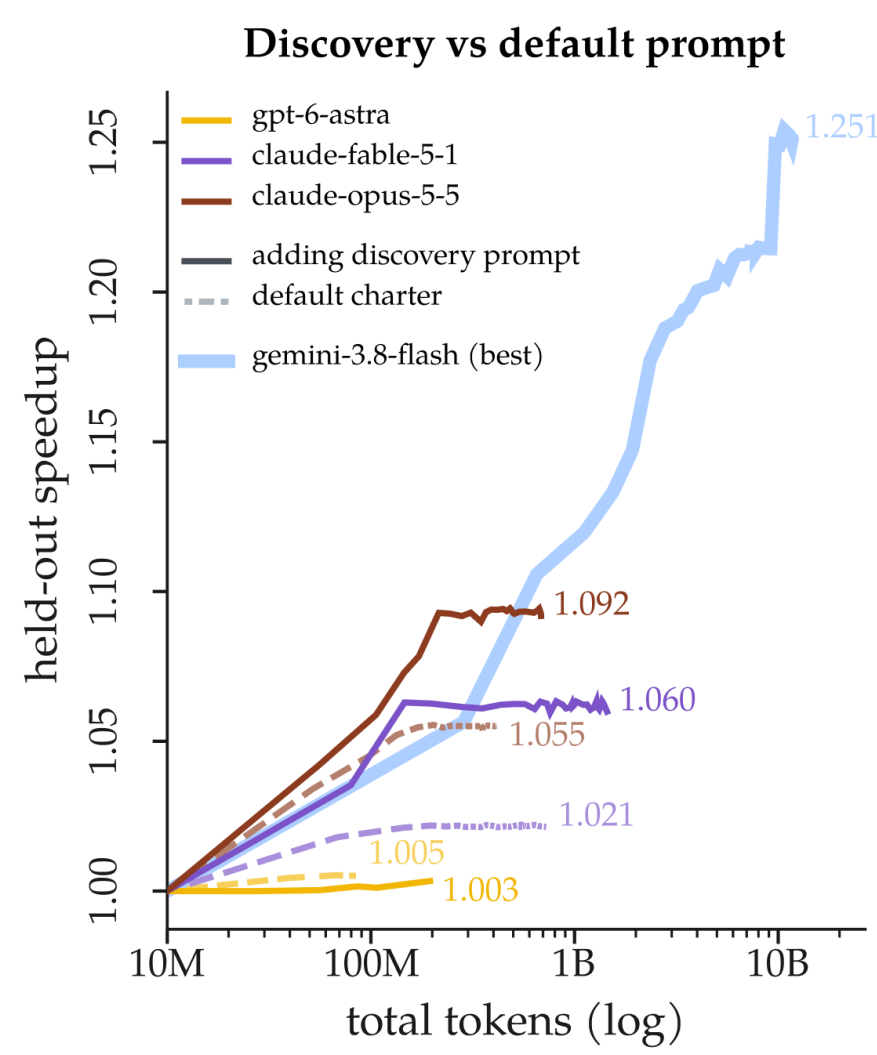

**Figure 4 Default and discovery charters.** Held-out speedup by total tokens (log scale) for three models with the default charter (dashed) and with a charter that asks agents to explore before concluding that no work remains (solid). The faded curve is Gemini's collaborative run with the default prompt.

**A few large improvements drive the overall gains without signs of hacking.** Looking more closely at the performance curves, we find that progress often comes from substantial speedups on individual held-out tasks. For example, Gemini's changes in World-1 make `cachecontrol`'s cache-hit task run 4.5× faster, even though the agents never see that package. The jumps in the Gemini and Kimi curves in World-2 similarly reflect 25–50% improvements in one or two held-out tasks (Fig. 3). We review the agents' traces, logs, and code changes and find no evidence that they search for the held-out packages or the evaluation setup. The improvements they find in their own packages thus help software beyond what they can see.

**Agents can improve speed at the expense of correctness.** While optimizing their packages, agents sometimes make changes that cause tests in dependent packages to fail. For example, `kimi-k2.7-code`'s changes repeatedly cause held-out tests to fail in World-3. Our main results use *carry* scoring, which retains earlier gains when tests fail. To see how much the reported gains depend on earlier passing results, we also apply *strict* scoring, where failing packages contribute no speedup. Table 1 shows cases where these two metrics differ: the strict policy lowers Gemini's final score from 1.144 to 1.101 in World-2, and from 1.305 to 1.294 in World-3, whereas it removes half of `kimi-k2.7-code`'s improvement.

**Table 1 Comparing held-out speedup under the strict or carry policy.** $S_{strict}$ shows when agents break the code.

| | $S_{carry}$ | $S_{strict}$ |
|---|---|---|
| **World-2** | | |
| Gemini | 1.144 | 1.101 (−.043) |
| **World-3** | | |
| Gemini | 1.305 | 1.294 (−.011) |
| Kimi | 1.124 | 1.065 (−.059) |

**Agents publish small releases with little code review.** Agent releases contain a median of 2 merged pull requests and 62 changed lines, compared with 12–15 pull requests and 524–608 lines for humans. In the runs we audit, agents rarely review one another's code before releasing it. We also find unintended changes and pull-request descriptions that overstate performance gains.

[5] These are the figure's scores under the carry rule. Both `claude-fable-5-1` runs fail the final correctness checks for `cachecontrol`, so its contribution retains the last passing measurements.

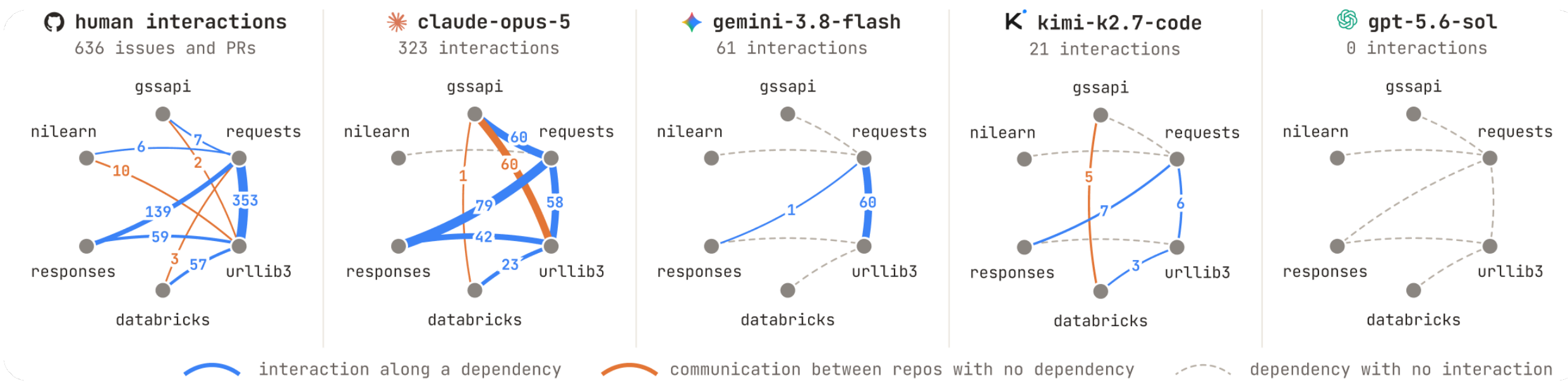

**Figure 5 Interactions between the six packages of World-1.** Left: human issues and pull requests on GitHub (636). Right: agent interactions, counting issues, pull requests, and comments between packages, for `claude-opus-5` (323), `gemini-3.8-flash` (61), `kimi-k2.7-code` (21), and `gpt-5.6-sol` (0). Numbers on edges are interaction counts.

These results show that agents can make substantial improvements to real software, but do not consistently make good use of the freedom to choose their own work. We next examine how agents' performance and behavior change when we vary whether they can communicate, which packages they maintain, and how they share resources within the world.

## 4 Communication Can Help Agents Discover Work

The promise of communication is that one agent's discovery can help another make progress (Park et al., 2026). A maintainer may notice that another package is slowing down its own code and share measurements with the other maintainer to help find a fix. Yet agents may exchange messages without using them effectively (Khatua et al., 2026). With one maintainer assigned to each of the six packages in World-1, we study whether communication helps agents discover useful work and improve downstream efficiency, and whether encouraging more communication leads to more gains.

**Experiment setup.** We first compare worlds where agents can communicate with one another (*connected*) against worlds where agents work independently (*isolated*). In connected runs, maintainers can inspect one another's repositories and exchange feedback through issues, pull requests, and reviews. Communication is *optional*, so agents are free to choose whether, when, and with whom to interact. In isolated runs, repositories remain private to their maintainer.

**Connected worlds outperform isolated ones for Gemini and Opus.** Gemini and Opus finish with higher held-out scores in their connected runs, while the differences for Sol, Kimi, and Astra are within the observed run-to-run variation (Fig. 1, right). Fig. 6 plots the same results against token counts. Over similar numbers of sessions, the connected Gemini and Opus runs spend 2.7 and 2.3 times as many tokens as their corresponding isolated runs, respectively. At equal token counts, the two settings reach similar scores, but connected agents find additional work and spend more tokens. In the isolated runs, Gemini and Opus agents report about half of their sessions as idle (53% and 48%), compared with at most 1% in connected runs.

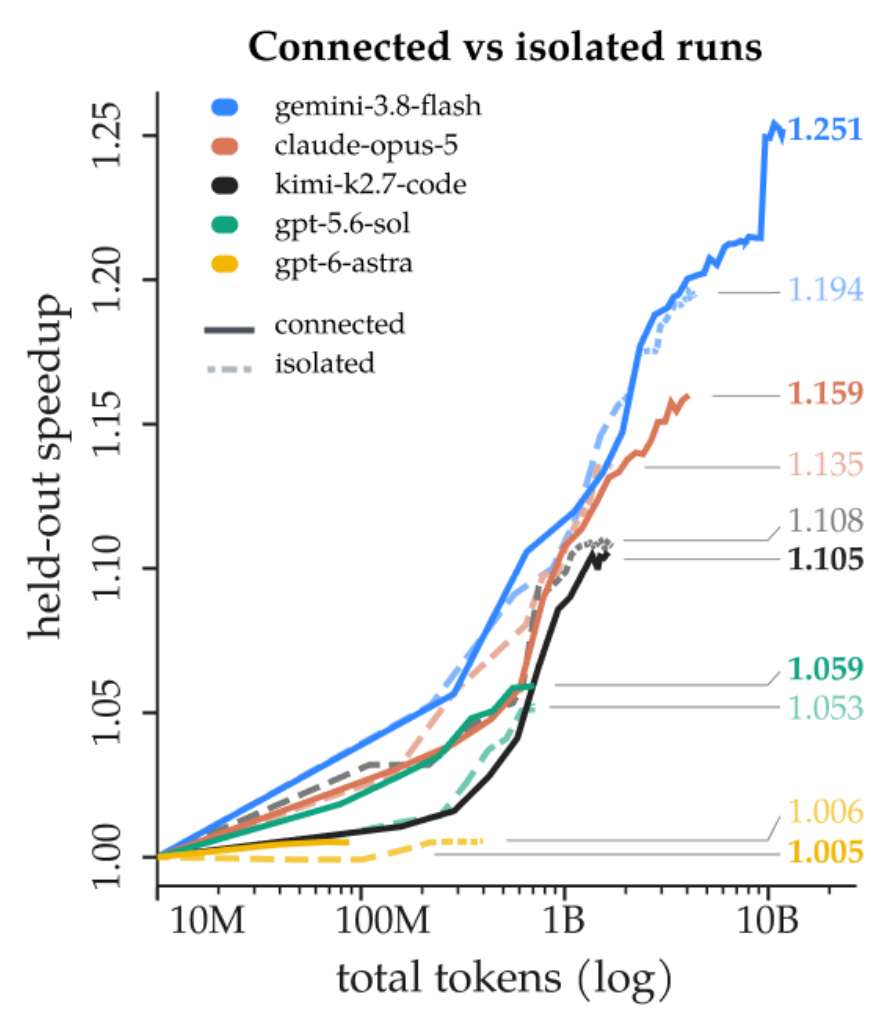

**Figure 6 Connected vs isolated runs.** We plot the same results from Fig. 1 (right) against token counts.

**Feedback from other maintainers can help an agent discover what to work on.** To see what information maintainers gain from one another, we inspect exchanges that lead to code changes and compare them with the isolated runs. In the Opus connected run, the `requests` maintainer identifies repeated work inside `urllib3` that `urllib3`'s benchmarks do not cover, and shares measurements that lead to an optimization. In the Gemini connected run, the `responses` maintainer catches a bug introduced by a `requests` optimization, even though the package's own tests pass. Its report helps the `requests` maintainer fix the bug while keeping the speedup. The isolated runs miss the `urllib3` optimization and the `requests` bug fix. See Appendix D.1 for details.

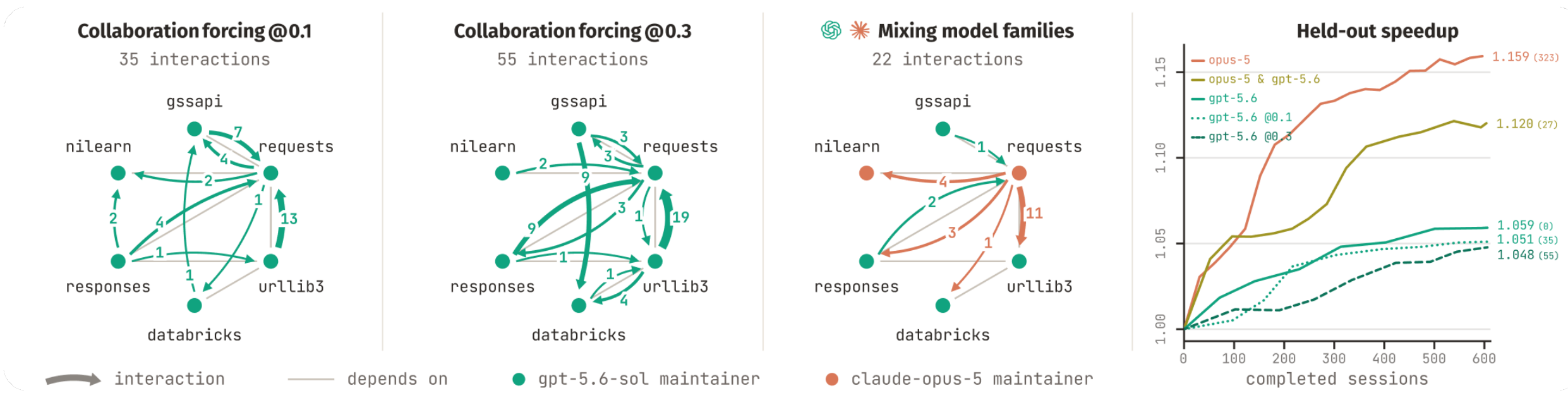

**Figure 7 Communication-forcing and mixed-model communication.** Prompting increases Sol's communication without improving performance, while mixing Sol with Opus leads to exchanges that Sol does not initiate on its own. The first two graphs show interactions when we prompt Sol agents to contact another maintainer in at least 10% or 30% of work sessions. The third shows interactions in a mixed world of four Sol and two Opus maintainers. The fourth compares their held-out scores over completed work sessions with those of all-Sol, Sol-Opus-mixed, and all-Opus worlds under optional communication.

**Agents share communication patterns with human maintainers.** We compare agents' communication patterns with human interactions across the same packages on GitHub (Fig. 5). Human interactions are most frequent between `requests` and `urllib3`. Gemini follows this pattern, with 60 of its 61 interactions between these two packages. Opus communicates between a wider range of packages, including pairs without a direct dependency. Kimi communicates less often, but also exchanges information between `requests` and `urllib3`, and between `requests` and `responses`. Sol makes no contact when communication is optional.

**More communication does not necessarily improve performance.** Opus has more than five times as many interactions as Gemini (Fig. 5), but Gemini achieves a higher held-out score (Fig. 1, right). To test whether increasing communication helps while keeping the model fixed, we focus on Sol. We run two *forced* variants with Sol: one asks agents to contact another maintainer in at least one of every ten work sessions, and the other raises this to three of every ten. The share of sessions involving contact rises from zero to 5.5% and 7.6%, respectively, but held-out performance does not improve. At comparable API cost, the one-in-ten variant scores 1.051 and the three-in-ten variant scores 1.048, compared with 1.059 for Sol under *optional* communication.

**Proactive agents can draw others into communication.** We next test whether Sol participates when more communicative agents initiate exchanges. In the *mixed* condition, four Sol maintainers work alongside two Opus maintainers, where communication remains *optional*. The mixed world scores 1.120, between the all-Sol world's 1.059 and the all-Opus world's 1.159 (Fig. 7).[6] Opus initiates all 12 cross-repository issue threads. Sol initiates none, but responds to 10 of the 11 issues addressed to its maintainers. These responses can lead to useful changes. In one exchange, Sol fixes an `urllib3` regression exposed by tests from an Opus maintainer.

These results show both the value and limits of agent communication. Exchanges between maintainers can reveal improvements and bugs they miss alone, but models differ in how they use this opportunity, and more communication does not guarantee better results. Sol makes no contact when communication is optional. Encouraging it to communicate increases exchanges without improving performance, while pairing it with more communicative Opus agents leads it to respond and contribute useful fixes.

[6] Introducing Opus changes communication and model capabilities, so this comparison does not isolate the effect of communication.

## 5 Clear Ownership Helps Agents Work Together

So far, we have divided the work by assigning each agent a package to maintain. Without these assignments, agents can choose which packages to improve, but they may duplicate work or leave some packages unattended. Can they divide the work effectively themselves, and does adding more agents help? We compare assigned and unassigned ownership, and vary the number of agents free to work on any package.

**Assigning each maintainer a package improves performance.** We first compare two runs with six Opus maintainers each (Fig. 8(a)). In one run, we assign each maintainer a package (*6 agents with assigned repos*). In the other, all six agents are free to modify all six packages (*without assigned repos - 6 agents*). At matched API cost, the first run achieves higher held-out scores than the second. Without assigned responsibilities, agents sometimes end up doing the same work. For example, two maintainers prepared the same `nilearn` release around the same time, and one had to discard most of its changes because the other finished first. While the agents eventually reached an agreement, they had already spent effort repeating each other's work.

**Without assigned ownership, multiple maintainers perform worse than one.** Without assignments, adding maintainers could spread useful work across more agents, but also create more overlap. We next compare one, four, six, and ten Opus maintainers without assigned packages (Fig. 8(a)). One maintainer outperforms all three multi-agent runs at matched API cost, with performance initially declining each time we add more agents. Additional maintainers spend some of their effort repeating or reviewing work that others have already done. In the ten-agent run, three agents independently rewrite the same twenty lines within three minutes, while a fourth duplicates a neighboring change. Reviews sometimes catch bugs, but also require additional work sessions. In the multi-agent runs, each merged change costs roughly two to three times as much in API usage as in the single-agent run. See Appendix D.2 for more examples.

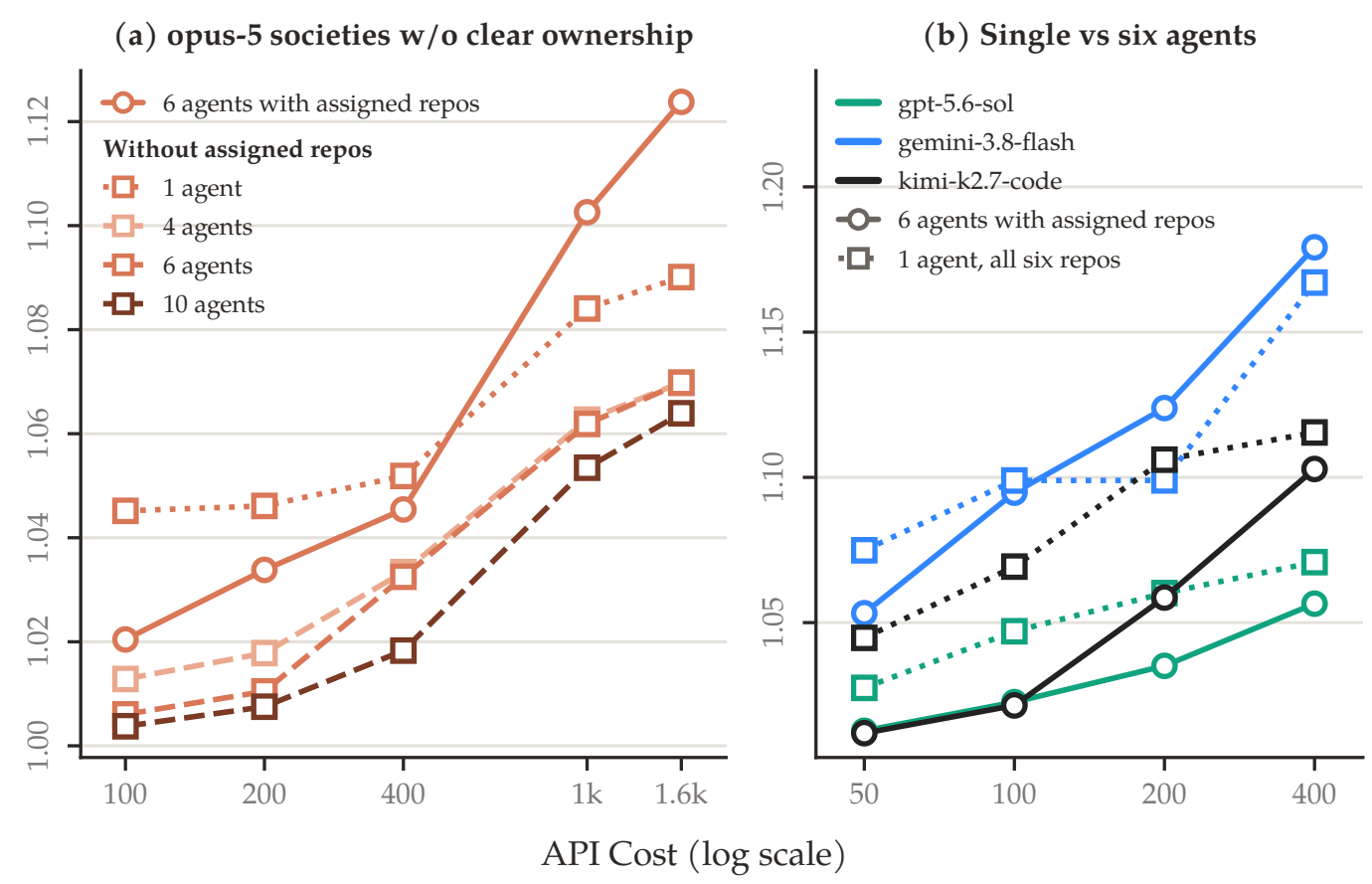

**Figure 8 Ownership and number of maintainers at matched API cost.** (a) Six assigned Opus maintainers outperform six without assigned packages. Adding unassigned maintainers provides no consistent gain. (b) One maintainer for all six packages versus six with one package each. One maintainer leads for Sol and Kimi; six Gemini maintainers lead at $200 and $400.

**Even with assigned ownership, more maintainers do not always help.** With assigned packages, six Opus maintainers outperform one agent maintaining all packages (Fig. 8(a)). We examine whether this trend extends to Sol, Kimi, and Gemini, comparing one maintainer responsible for all six packages with six maintainers assigned one package each (Fig. 8(b)). For Sol and Kimi, one maintainer consistently achieves higher held-out scores than six. The Gemini results are less conclusive, with one maintainer leading at lower spending and six at higher spending. Opus' success with six maintainers might be related to Opus also initiating the most communication among these models (Fig. 5). It is possible that its willingness to initiate exchanges helps it benefit from distributing work across maintainers.

Letting agents choose their responsibilities makes dividing the work part of the task. Agents can develop their own coordination practices, but duplication and additional review can make their work more costly. Assigning packages provides a division of responsibility while preserving freedom to choose useful improvements. Whether distributing those responsibilities across agents improves performance still depends on the model.

## 6 Agents Struggle to Effectively Allocate Shared Resources

Assigning each maintainer a package leaves another decision unresolved: how much effort should each package receive? So far, every maintainer has received 100 work sessions, regardless of the opportunities it discovers. To investigate the effect of this choice, we consider a shared budget setting. A shared budget allows maintainers to request more resources for promising work, but requires them to judge one another's proposals and decide which work to support. We study how Sol, Opus, and Gemini make these decisions in World-1, keeping one assigned maintainer per package.

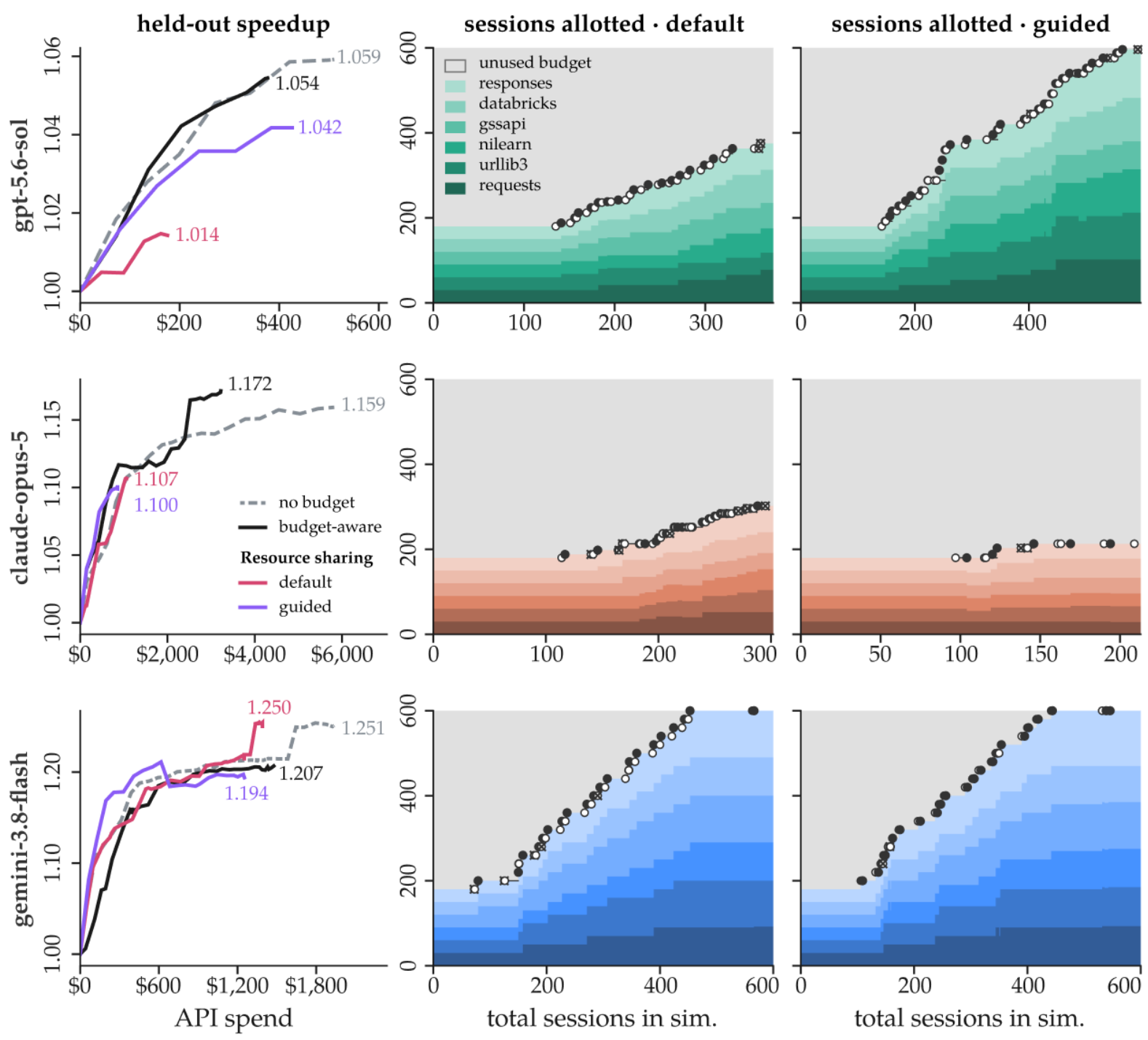

**Figure 9 Held-out speedup and session allocation with a shared budget.** Left: held-out speedup by API spend under different conditions. Middle and right: sessions allotted to each agent (stacked shades) and the unused shared budget (gray). Each maintainer starts with 30 sessions, and the world has 600 in total. Markers are budget proposals, hollow when filed, filled when merged, and crossed when closed without merging.

**Experiment setup.** Each of the six maintainers starts with 30 sessions and can request more from a shared pool of 420. This gives the world 600 sessions in total, the same as with individual budgets of 100. Maintainers can also donate unused sessions from their own budget to another maintainer. Only maintainers with sessions left can vote. A request needs approval from a majority of the other maintainers, and the pool closes when fewer than two can vote. We keep these rules fixed and compare two versions of the instructions. The *default* instructions explain how to request, approve, and donate sessions, and ask maintainers to judge whether the proposed work is worth the sessions. The *guided* instructions further ask maintainers to prioritize shared resources for the work most useful to the world as a whole, to weigh their own needs, and to consider the cost of reviewing many small requests. Appendix B details the prompts used for this experiment.[7]

**Agents actively participate in sharing the budget.** Under the default instructions, agents request sessions, vote, or discuss the budget in 19% of Sol's completed sessions, 49% of Opus's, and 15% of Gemini's (Table 2).[8] Agents approve nearly every request they vote on. Sol casts 53 votes to approve requests and none to reject them, while Opus casts 82 to approve and six to reject.[9] Agents also usually approve requests under the guided instructions. Some maintainers go further and donate their own sessions so others can keep working. For example, a guided Opus maintainer donates four sessions after identifying a possible optimization in another maintainer's package. Appendix D.3 reports the donation counts and further examples.

**Active participation does not always lead to effective resource use.** Despite this active participation, the default Sol and Opus runs stop with 225 and 298 shared sessions unused, respectively (Fig. 9, middle). For requests that receive shared sessions, Sol and Opus have a median of only one and three sessions left when they ask, respectively (Table 2). As maintainers run out of sessions, they lose the ability to vote, and both runs eventually have too few eligible voters to keep the pool open. Both finish below the individual-budget runs with undisclosed budgets, with scores of 1.014 versus 1.059 for Sol and 1.107 versus 1.159 for Opus (Fig. 9, left). In contrast, Gemini distributes all 420 shared sessions under the default instructions. Its successful requests arrive while maintainers still have a median of nine sessions left. It also asks for more sessions at once, a median of 20 compared with 12 for Sol and eight for Opus (Table 2). Its final score of 1.250 nearly matches the score with individual budgets (1.251), showing that agents can use a shared budget without sacrificing final performance in this comparison. See Appendix C for details on budget implementation.

[7] Each condition has one run. Session limits are 30 minutes for Sol and 60 minutes for Opus. The Sol and Opus comparisons also use different clusters for the *default* and *guided* runs, and the *guided* Sol run used subscription access rather than API billing, so the comparison does not isolate the change in instructions.

[8] The activity records cover actions on the budget board and package repositories, but exclude local code edits and tests. These percentages do not measure how much of each session agents spend on budgeting.

[9] Each pull request on the budget board counts as one proposal, whether it requests shared sessions or donates sessions to another maintainer. We count replacements as separate proposals and count a proposal as completed when it merges. Vote counts include repeated votes and votes that maintainers later change or withdraw.

**Table 2 Budget use under default and guided instructions.** Request statistics include only receiving sessions from the shared pool. Remaining sessions refer to the agent's unstarted sessions when submitting the request. Percentages show the share of completed work sessions involving budget request activities.

| | Default | | | Guided | | |
|---|---|---|---|---|---|---|
| | Sol | Opus | Gemini | Sol | Opus | Gemini |
| Total completed work sessions | 372 | 277 | 493 | 594 | 213 | 596 |
| Own sessions remaining when requesting more (median) | 1 | 3 | 9 | 4 | 4 | 31 |
| Sessions requested per request (median) | 12 | 8 | 20 | 12 | 10 | 20 |
| Shared sessions left unused (out of 420) | 225 | 298 | 0 | 4 | 387 | 0 |
| Sessions involving budget requests, votes, or discussion | 19% | 49% | 15% | 28% | 20% | 15% |

**Guidance changes how agents share sessions, but does not consistently improve performance.** The default runs differ sharply in how much of the pool they use. We examine whether asking maintainers to prioritize work that benefits the whole world leads to more effective allocation. Budget activity appears in 28% of Sol's completed sessions under guidance, up from 19% under the default instructions (Table 2). It draws 416 of the 420 shared sessions, compared with 195, sometimes requesting 24 at once to reduce repeated reviews. Its final score rises from 1.014 to 1.042, but remains below the individual-budget baseline. Opus moves in the opposite direction, with budget activity falling from 49% to 20% of completed sessions and the agents drawing only 33 shared sessions rather than 122. It finishes at 1.100, slightly below its default score of 1.107. Gemini uses the full pool in both runs, with budget activity in about 15% of completed sessions under either set of instructions. In both runs, all 420 shared sessions have been allocated by the time about 450 agent sessions have started (Fig. 9, bottom row). Its guided run gives `responses` 130 sessions and `requests-gssapi` 110, leaving the other maintainers with 85–93 each after donations. Despite using the full pool, it finishes at 1.194, below its default score of 1.250.[10]

**Agents can fund work that does not improve held-out performance.** Beyond deciding how much to request, maintainers must judge which work deserves support. In the guided Opus run, the `databricks-sql-connector` maintainer requests more sessions after reporting 12 releases with measured speedups and concrete plans for further work. Three maintainers approve the request, explicitly citing its results and plans. Databricks receives 15 of the 33 sessions allocated from the shared pool in that run. Improvements specific to Databricks do not directly raise the held-out score, which measures changes to `requests` and `urllib3`. Their decision rests on reported improvements within a package, which need not translate into gains on unseen dependent packages.

With a shared budget, whether an agent can keep working depends on decisions by other maintainers. Agents readily support one another, yet resources can still go unused, and the work they fund may not benefit held-out packages. Making good use of a shared budget takes more than a willingness to share.

## 7 Related Work

**Multi-agent collaboration.** Multi-agent systems divide work through assigned roles (Hong et al., 2024; Qian et al., 2023; Huang et al., 2023), task decomposition (Chen et al., 2024a; Geng and Neubig, 2026; Anthropic,

[10]The guided Gemini run also introduces a public-interface regression affecting `requests-cache`, which contributes a score of 1.0 under the internals-aware rule. The performance difference therefore cannot be attributed to allocation decisions alone.

2025), and workflows that adapt to feedback (Hu et al., 2024). Communication can improve collective search (Park et al., 2026), but agents can struggle to coordinate code changes or combine information (Khatua et al., 2026; Li et al., 2026b). The benefits of adding agents depend on the task and model (Kim et al., 2025), while studies of self-organization examine how models and protocols affect collaboration (Dochkina, 2026) and how agents learn coordination strategies that transfer to new problems (Pappu et al., 2026). Evaluations also study interaction protocols (Agashe et al., 2025; Zhu et al., 2025a), social behavior (Zhou et al., 2024), and collaboration processes and failures (Chen et al., 2026b; Cemri et al., 2025). Persistent environments also support self-organization (Chen et al., 2024b; Tessera et al., 2026) and simulated societies that evolve through agents' actions (Park et al., 2023; Piao et al., 2025; Altera.AL et al., 2024). Shared-resource simulations examine whether agents sustain a common resource (Piatti et al., 2024), and work on agent economies proposes ways to organize exchanges and allocate resources (Tomasev et al., 2025). We study collaboration through ongoing maintenance of real software, where agents discover tasks themselves and we vary communication, ownership, and resource allocation to examine how they work together.

**Software engineering benchmarks.** Coding benchmarks evaluate assigned issue resolution (Jimenez et al., 2024; Deng et al., 2025), work across repositories (Chen et al., 2026a), and maintenance through successive changes (Le et al., 2025; Lam et al., 2026; Chen et al., 2026c). Work on bug discovery without issue reports gives agents more responsibility for finding tasks (Li et al., 2026a; Bui and Evangelopoulos, 2026). Performance optimization benchmarks ask agents to make real software faster while preserving correctness (Shetty et al., 2025; He et al., 2025; Ma et al., 2025; Sehgal et al., 2026; Nangia et al., 2026). Our setting combines this optimization goal with task discovery and collaboration across interdependent packages. Evaluation studies also identify weak tests that accept incorrect patches, issues that reveal solutions, and other flaws in agent benchmarks (Aleithan et al., 2024; Yu et al., 2025; Wang et al., 2025; Zhu et al., 2025b). Human review and continuously updated tasks address some of these problems (OpenAI, 2024; Zhang et al., 2025). We test whether agents' improvements benefit dependent packages using human-authored tests and benchmarks that agents cannot inspect or modify. This measures generalization beyond the work agents choose for themselves. We also use instruction counts to reduce the sensitivity to execution conditions that affects performance measurements (Chen et al., 2026d).

**Coordination in open-source software.** Software dependencies connect maintainers' work, so a change in one package can require changes in others (Malone and Crowston, 1994; Kikas et al., 2017; Decan et al., 2019; Venturini et al., 2023). Maintainers must negotiate who takes responsibility for these changes and the effort they require (Bogart et al., 2016, 2021). Studies link communication that follows work dependencies to faster problem resolution (Cataldo et al., 2006, 2008), and code ownership and contributor expertise to how often software defects occur (Bird et al., 2011). Brooks (1975) describes how adding developers can increase training and communication costs. These findings motivate our experiments on how agents communicate, divide responsibilities, and spend their limited resources while maintaining interdependent software.

## 8 Conclusion

We introduce Software World, a framework for studying collaboration among agents that choose their own work while maintaining real, interdependent software. We vary their working conditions and evaluate whether their changes improve software in dependent packages they never see. Our experiments show that agents can discover substantial improvements, yet some frontier models stop searching while useful work remains. Feedback from other maintainers can help agents find optimizations and catch bugs they miss alone. But more communication does not reliably improve performance, and additional maintainers can spend effort repeating one another's work. Clear package responsibilities help Opus agents work together, while shared budgets reveal that agents can actively support one another yet leave resources unused. We release the framework and traces as a foundation for studying when agents benefit from working together and how to make that collaboration more effective.

**Acknowledgments** We thank Lucas Torroba Hennigen, John Giorgi, William Shen, Alex Gu, Carolyn Zou, and Scott Yih for helpful discussions, and OpenAI and Fireworks.AI for providing API credits. SW was supported in part by ONR N000142312840 and gifts from Microsoft AI and Cisco Research. MI was supported by a Google Cloud Credit award that enabled Gemini experiments in this work.

## References

Saaket Agashe, Yue Fan, Anthony Reyna, and Xin Eric Wang. LLM-coordination: Evaluating and analyzing multi-agent coordination abilities in large language models. In *Findings of the Association for Computational Linguistics: NAACL 2025*, pages 8053–8072, Albuquerque, New Mexico, 2025. Association for Computational Linguistics. doi: 10.18653/v1/2025.findings-naacl.448. URL `https://aclanthology.org/2025.findings-naacl.448/`.

Reem Aleithan, Haoran Xue, Mohammad Mahdi Mohajer, Elijah Nnorom, Gias Uddin, and Song Wang. SWE-Bench+: Enhanced coding benchmark for LLMs. *arXiv preprint arXiv:2410.06992*, 2024. URL `https://arxiv.org/abs/2410.06992`.

Altera.AL et al. Project Sid: Many-agent simulations toward AI civilization. *arXiv preprint arXiv:2411.00114*, 2024.

Anthropic. How we built our multi-agent research system. `https://www.anthropic.com/engineering/multi-agent-research-system`, 2025. Published June 13, 2025; accessed September 25, 2026.

Anthropic. Formalizing Fermat's last theorem. `https://www.anthropic.com/research/formalizing-fermats-last-theorem`, 2026. September 4, 2026, accessed 2026-09-26.

Christian Bird, Nachiappan Nagappan, Brendan Murphy, Harald Gall, and Premkumar Devanbu. Don't touch my code!: Examining the effects of ownership on software quality. In *Proceedings of the 19th ACM SIGSOFT Symposium and the 13th European Conference on Foundations of Software Engineering*, pages 4–14, 2011. doi: 10.1145/2025113.2025119.

Vincent D Blondel, Jean-Loup Guillaume, Renaud Lambiotte, and Etienne Lefebvre. Fast unfolding of communities in large networks. *Journal of statistical mechanics: theory and experiment*, 2008(10):P10008, 2008.

Christopher Bogart, Christian Kästner, James Herbsleb, and Ferdian Thung. How to break an API: Cost negotiation and community values in three software ecosystems. In *Proceedings of the 24th ACM SIGSOFT International Symposium on Foundations of Software Engineering (FSE)*, pages 109–120, 2016.

Christopher Bogart, Christian Kästner, James Herbsleb, and Ferdian Thung. When and how to make breaking changes: Policies and practices in 18 open source software ecosystems. *ACM Transactions on Software Engineering and Methodology*, 30(4):1–56, 2021.

Michael E. Bratman. Shared cooperative activity. *The Philosophical Review*, 101(2):327–341, 1992. doi: 10.2307/2185537.

Frederick P Brooks. *The Mythical Man-Month: Essays on Software Engineering*. Addison-Wesley, 1975.

Nghi D. Q. Bui and Georgios Evangelopoulos. Agentic coding needs proactivity, not just autonomy. *arXiv preprint arXiv:2605.06717*, 2026. URL `https://arxiv.org/abs/2605.06717`.

Nicholas Carlini. Building a C compiler with a team of parallel Claudes. `https://www.anthropic.com/engineering/building-c-compiler`, February 2026. Published February 5, 2026; accessed October 1, 2026.

Marcelo Cataldo, Patrick A Wagstrom, James D Herbsleb, and Kathleen M Carley. Identifying coordination requirements in software development: implications for coordination requirements. In *Proceedings of the ACM Conference on Computer Supported Cooperative Work (CSCW)*, pages 353–362, 2006.

Marcelo Cataldo, James D Herbsleb, and Kathleen M Carley. Socio-technical congruence: a framework for assessing the impact of technical and work dependencies on software development productivity. In *Proceedings of the ACM-IEEE International Symposium on Empirical Software Engineering and Measurement (ESEM)*, pages 2–11, 2008.

Mert Cemri, Melissa Z. Pan, Shuyi Yang, Lakshya A. Agrawal, Bhavya Chopra, Rishabh Tiwari, Kurt Keutzer, Aditya Parameswaran, Dan Klein, Kannan Ramchandran, Matei A. Zaharia, Joseph E. Gonzalez, and Ion Stoica. Why do multi-agent LLM systems fail? In *Advances in Neural Information Processing Systems*, volume 38, 2025. doi: 10.52202/085713-4082. URL https://proceedings.neurips.cc/paper_files/paper/2025/hash/b1041e52d3be19f0a9bc491657488e4a-Abstract-Datasets_and_Benchmarks_Track.html.

Dong Chen, Shaoxin Lin, Muhan Zeng, Daoguang Zan, Jian-Gang Wang, Anton Cheshkov, Jun Sun, Hao Yu, Guoliang Dong, Artem Aliev, Jie Wang, Xiao Cheng, Guangtai Liang, Yuchi Ma, Pan Bian, Tao Xie, and Qianxiang Wang. CodeR: Issue resolving with multi-agent and task graphs. *arXiv preprint arXiv:2406.01304*, 2024a. URL https://arxiv.org/abs/2406.01304.

Guoxin Chen, Fanzhe Meng, Jiale Zhao, Minghao Li, Daixuan Cheng, Huatong Song, Jie Chen, Yuzhi Lin, Hui Chen, Xin Zhao, Ruihua Song, Chang Liu, Cheng Chen, Kai Jia, and Ji-Rong Wen. BeyondSWE: Can current code agent survive beyond single-repo bug fixing? *arXiv preprint arXiv:2603.03194*, 2026a. URL https://arxiv.org/abs/2603.03194.

Jiaju Chen, Bo Sun, Yuxuan Lu, Yun Wang, Dakuo Wang, and Bingsheng Yao. CollabSim: A CSCW-grounded methodology for investigating collaborative competence of LLM agents through controlled multi-agent experiments. *arXiv preprint arXiv:2606.06399*, 2026b.

Jialong Chen, Xander Xu, Hu Wei, Chuan Chen, and Bing Zhao. SWE-CI: Evaluating agent capabilities in maintaining codebases via continuous integration. *arXiv preprint arXiv:2603.03823*, 2026c. URL https://arxiv.org/abs/2603.03823.

Jiaqi Chen, Yuxian Jiang, Jiachen Lu, and Li Zhang. S-agents: Self-organizing agents in open-ended environments. *CoRR*, abs/2402.04578, 2024b. doi: 10.48550/arXiv.2402.04578. URL https://arxiv.org/abs/2402.04578.

Zhi Chen, Zhensu Sun, Yuling Shi, David Lo, and Lingxiao Jiang. Are performance-optimization benchmarks reliably measuring coding agents? *arXiv preprint arXiv:2607.01211*, 2026d. URL https://arxiv.org/abs/2607.01211.

Alexandre Decan, Tom Mens, and Philippe Grosjean. An empirical comparison of dependency network evolution in seven software packaging ecosystems. *Empirical Software Engineering*, 24(1):381–416, 2019.

Xiang Deng, Jeff Da, Edwin Pan, Yannis Yiming He, Charles Ide, Kanak Garg, Niklas Lauffer, Andrew Park, Nitin Pasari, Chetan Rane, Karmini Sampath, Maya Krishnan, Srivatsa Kundurthy, Sean Hendryx, Zifan Wang, Vijay Bharadwaj, Jeff Holm, Raja Aluri, Chen Bo Calvin Zhang, Noah Jacobson, Bing Liu, and Brad Kenstler. SWE-Bench Pro: Can AI agents solve long-horizon software engineering tasks? *arXiv preprint arXiv:2509.16941*, 2025. URL https://arxiv.org/abs/2509.16941.

Victoria Dochkina. Drop the hierarchy and roles: How self-organizing LLM agents outperform designed structures. *arXiv preprint arXiv:2603.28990*, 2026.

Philip J. Fleming and John J. Wallace. How not to lie with statistics: the correct way to summarize benchmark results. *Communications of the ACM*, 29(3):218–221, 1986.

Adam Fourney, Gagan Bansal, Hussein Mozannar, Cheng Tan, Eduardo Salinas, Friederike Niedtner, Grace Proebsting, Griffin Bassman, Jack Gerrits, Jacob Alber, et al. Magentic-one: A generalist multi-agent system for solving complex tasks. *arXiv preprint arXiv:2411.04468*, 2024.

Jiayi Geng and Graham Neubig. Effective strategies for asynchronous software engineering agents. *arXiv preprint arXiv:2603.21489*, 2026. URL https://arxiv.org/abs/2603.21489.

Harbor Framework Team. Harbor: A framework for evaluating and optimizing agents and models in container environments, 2026. URL https://doi.org/10.5281/zenodo.20953922.

Xinyi He, Qian Liu, Mingzhe Du, Lin Yan, Zhijie Fan, Yiming Huang, Yin Zheng, Zejian Yuan, and Zejun Ma. SWE-Perf: Can language models optimize code performance on real-world repositories? *arXiv preprint arXiv:2507.12415*, 2025. URL https://arxiv.org/abs/2507.12415.

Sirui Hong, Mingchen Zhuge, Jiaqi Chen, Xiawu Zheng, Yuheng Cheng, Ceyao Zhang, Jinlin Wang, Zili Wang, Steven Ka Shing Yau, Zijuan Lin, Liyang Zhou, Chenyu Ran, Lingfeng Xiao, Chenglin Wu, and Jürgen Schmidhuber. MetaGPT: Meta programming for a multi-agent collaborative framework. In *International Conference on Learning Representations (ICLR)*, 2024. URL `https://arxiv.org/abs/2308.00352`.

Yue Hu, Yuzhu Cai, Yaxin Du, Xinyu Zhu, Xiangrui Liu, Zijie Yu, Yuchen Hou, Shuo Tang, and Siheng Chen. Self-evolving multi-agent collaboration networks for software development. *arXiv preprint arXiv:2410.16946*, 2024. URL `https://arxiv.org/abs/2410.16946`.

Dong Huang, Jie M. Zhang, Michael Luck, Qingwen Bu, Yuhao Qing, and Heming Cui. AgentCoder: Multi-agent-based code generation with iterative testing and optimisation. *arXiv preprint arXiv:2312.13010*, 2023. URL `https://arxiv.org/abs/2312.13010`.

Carlos E. Jimenez, John Yang, Alexander Wettig, Shunyu Yao, Kexin Pei, Ofir Press, and Karthik Narasimhan. SWE-bench: Can language models resolve real-world GitHub issues? In *International Conference on Learning Representations (ICLR)*, 2024. URL `https://arxiv.org/abs/2310.06770`.

Arpandeep Khatua, Hao Zhu, Peter Tran, Arya Prabhudesai, Frederic Sadrieh, Johann K. Lieberwirth, Xinkai Yu, Yicheng Fu, Michael J. Ryan, Jiaxin Pei, and Diyi Yang. CooperBench: Why coding agents cannot be your teammates yet. *arXiv preprint arXiv:2601.13295*, 2026. URL `https://arxiv.org/abs/2601.13295`.

Riivo Kikas, Georgios Gousios, Marlon Dumas, and Dietmar Pfahl. Structure and evolution of package dependency networks. In *Proceedings of the 14th International Conference on Mining Software Repositories (MSR)*, pages 102–112, 2017.

Yubin Kim et al. Towards a science of scaling agent systems. *arXiv preprint arXiv:2512.08296*, 2025.

Man Ho Lam, Chaozheng Wang, Hange Liu, Jingyu Xiao, Haau-sing Li, Jen-tse Huang, Terry Yue Zhuo, and Michael R. Lyu. SWE-Chain: Benchmarking coding agents on chained release-level package upgrades. *arXiv preprint arXiv:2605.14415*, 2026. URL `https://arxiv.org/abs/2605.14415`.

Tue Le, Minh V. T. Thai, Dung Nguyen Manh, Huy Phan Nhat, and Nghi D. Q. Bui. SWE-EVO: Benchmarking coding agents in long-horizon software evolution scenarios. *arXiv preprint arXiv:2512.18470*, 2025. URL `https://arxiv.org/abs/2512.18470`.

Haobin Li, Ping Deng, Weizhong Qian, Liang Jiang, Zhenyu Huang, Mouxing Yang, and Xi Peng. Active-SWE: Benchmarking coding agents for proactive bug fixing without issue reports. *arXiv preprint arXiv:2608.04682*, 2026a. URL `https://arxiv.org/abs/2608.04682`.

Yuxuan Li, Aoi Naito, and Hirokazu Shirado. Systematic failures in collective reasoning under distributed information in multi-agent LLMs. In *International Conference on Machine Learning*, 2026b. URL `https://arxiv.org/abs/2505.11556`.

Jeffrey Jian Ma, Milad Hashemi, Amir Yazdanbakhsh, Kevin Swersky, Ofir Press, Enhui Li, Vijay Janapa Reddi, and Parthasarathy Ranganathan. SWE-fficiency: Can language models optimize real-world repositories on real workloads? *arXiv preprint arXiv:2511.06090*, 2025. URL `https://arxiv.org/abs/2511.06090`.

Thomas W Malone and Kevin Crowston. The interdisciplinary study of coordination. *ACM Computing Surveys*, 26(1):87–119, 1994.

Ayush Nangia, Shikhar Mishra, Aman Gokrani, and Paras Chopra. ISO-Bench: Can coding agents optimize real-world inference workloads? *arXiv preprint arXiv:2602.19594*, 2026. URL `https://arxiv.org/abs/2602.19594`.

OpenAI. Introducing SWE-bench verified. OpenAI blog post, August 2024. URL `https://openai.com/index/introducing-swe-bench-verified/`. Human-validated 500-instance subset of SWE-bench.

OpenAI. On the Navier–Stokes Millennium Prize Problem. `https://openai.com/index/navier-stokes-solution/`, September 2026. Published September 8, 2026; updated September 10, 2026; accessed October 1, 2026.

Melissa Z. Pan, Mert Cemri, Lakshya A. Agrawal, Shuyi Yang, Bhavya Chopra, Rishabh Tiwari, Kurt Keutzer, Aditya Parameswaran, Kannan Ramchandran, Dan Klein, Joseph E. Gonzalez, Matei Zaharia, and Ion Stoica. Why do multiagent systems fail? In *ICLR 2025 Workshop on Building Trust in Language Models and Applications*, 2025. URL `https://openreview.net/forum?id=wM521FqPvI`.

Aneesh Pappu, Mirac Suzgun, Yongchan Kwon, Federico Bianchi, Batu El, Mykel J. Kochenderfer, Hancheng Cao, and James Zou. Self-organizing agent teams learn to reason together. *arXiv preprint arXiv:2609.22682*, 2026.

Jongho Park, Vasilis Kontonis, Shivam Garg, Akshay Krishnamurthy, and Dimitris Papailiopoulos. Scaling discovery through test-time communication. *arXiv preprint arXiv:2609.21032*, 2026.

Joon Sung Park, Joseph C. O'Brien, Carrie J. Cai, Meredith Ringel Morris, Percy Liang, and Michael S. Bernstein. Generative agents: Interactive simulacra of human behavior. *arXiv preprint arXiv:2304.03442*, 2023. URL `https://arxiv.org/abs/2304.03442`.

Jinghua Piao, Yuwei Yan, Jun Zhang, Nian Li, Junbo Yan, Xiaochong Lan, Zhihong Lu, Zhiheng Zheng, Jing Yi Wang, Di Zhou, Chen Gao, Fengli Xu, Fang Zhang, Ke Rong, Jun Su, and Yong Li. AgentSociety: Large-scale simulation of LLM-driven generative agents advances understanding of human behaviors and society. *arXiv preprint arXiv:2502.08691*, 2025. URL `https://arxiv.org/abs/2502.08691`.

Giorgio Piatti, Zhijing Jin, Max Kleiman-Weiner, Bernhard Schölkopf, Mrinmaya Sachan, and Rada Mihalcea. Cooperate or collapse: Emergence of sustainable cooperation in a society of LLM agents. In *Advances in Neural Information Processing Systems*, 2024. URL `https://arxiv.org/abs/2404.16698`.

Chen Qian, Wei Liu, Hongzhang Liu, Nuo Chen, Yufan Dang, Jiahao Li, Cheng Yang, Weize Chen, Yusheng Su, Xin Cong, Juyuan Xu, Dahai Li, Zhiyuan Liu, and Maosong Sun. ChatDev: Communicative agents for software development. *arXiv preprint arXiv:2307.07924*, 2023. URL `https://arxiv.org/abs/2307.07924`.

Atharva Sehgal, James Hou, Akanksha Sarkar, Ishaan Mantripragada, Swarat Chaudhuri, Jennifer J. Sun, and Yisong Yue. FormulaCode: Evaluating agentic optimization on large codebases. *arXiv preprint arXiv:2603.16011*, 2026. URL `https://arxiv.org/abs/2603.16011`.

Rulin Shao, Shannon Zejiang Shen, Junjie Oscar Yin, Yuetai Li, Minheng Wang, Hamish Ivison, Radha Poovendran, Nathan Lambert, Teng Xiao, Mike Lewis, Wen-tau Yih, Luke Zettlemoyer, and Pang Wei Koh. Context language models. *arXiv preprint arXiv:2609.37725*, 2026.

Manish Shetty, Naman Jain, Jinjian Liu, Vijay Kethanaboyina, Koushik Sen, and Ion Stoica. GSO: Challenging software optimization tasks for evaluating SWE-agents. *arXiv preprint arXiv:2505.23671*, 2025. URL `https://arxiv.org/abs/2505.23671`.

Alexander Shypula, Aman Madaan, Yimeng Zeng, Uri Alon, Jacob Gardner, Milad Hashemi, Graham Neubig, Parthasarathy Ranganathan, Osbert Bastani, and Amir Yazdanbakhsh. Learning performance-improving code edits. In *International Conference on Learning Representations (ICLR)*, 2024.

Kale-ab Abebe Tessera, Andras Szecsenyi, Cameron Barker, Alexander Rutherford, Davide Paglieri, Aidan Scannell, Henry Gouk, Elliot J. Crowley, Tim Rocktäschel, and Amos J. Storkey. Benchmarking open-ended multi-agent coordination in language agents. *CoRR*, abs/2606.08340, 2026. doi: 10.48550/arXiv.2606.08340. URL `https://arxiv.org/abs/2606.08340`.

Nenad Tomasev, Matija Franklin, Joel Z Leibo, Julian Jacobs, William A Cunningham, Iason Gabriel, and Simon Osindero. Virtual agent economies. *arXiv preprint arXiv:2509.10147*, 2025.

Daniel Venturini, Filipe Roseiro Cogo, Ivanilton Polato, Marco A Gerosa, and Igor Scaliante Wiese. I depended on you and you broke me: An empirical study of manifesting breaking changes in client packages. *ACM Transactions on Software Engineering and Methodology*, 32(4):1–26, 2023.

You Wang, Michael Pradel, and Zhongxin Liu. Are "solved issues" in SWE-bench really solved correctly? An empirical study. *arXiv preprint arXiv:2503.15223*, 2025. URL `https://arxiv.org/abs/2503.15223`.

Boxi Yu, Yuxuan Zhu, Pinjia He, and Daniel Kang. UTBoost: Rigorous evaluation of coding agents on SWE-Bench. *arXiv preprint arXiv:2506.09289*, 2025. URL `https://arxiv.org/abs/2506.09289`.

Linghao Zhang, Shilin He, Chaoyun Zhang, Yu Kang, Bowen Li, Chengxing Xie, Junhao Wang, Maoquan Wang, Yufan Huang, Shengyu Fu, Elsie Nallipogu, Qingwei Lin, Yingnong Dang, Saravan Rajmohan, and Dongmei Zhang. SWE-bench goes live! *arXiv preprint arXiv:2505.23419*, 2025. URL `https://arxiv.org/abs/2505.23419`.

Xuhui Zhou, Hao Zhu, Leena Mathur, Ruohong Zhang, Haofei Yu, Zhengyang Qi, Louis-Philippe Morency, Yonatan Bisk, Daniel Fried, Graham Neubig, and Maarten Sap. SOTOPIA: Interactive evaluation for social intelligence in language agents. In *International Conference on Learning Representations*, 2024. URL `https://arxiv.org/abs/2310.11667`.

Kunlun Zhu, Hongyi Du, Zhaochen Hong, Xiaocheng Yang, Shuyi Guo, Zhe Wang, Zhenhailong Wang, Cheng Qian, Xiangru Tang, Heng Ji, and Jiaxuan You. MultiAgentBench: Evaluating the collaboration and competition of LLM agents. In *Proceedings of the 63rd Annual Meeting of the Association for Computational Linguistics (Volume 1: Long Papers)*, pages 8580–8622, Vienna, Austria, 2025a. Association for Computational Linguistics. doi: 10.18653/v1/2025.acl-long.421. URL `https://aclanthology.org/2025.acl-long.421/`.

Yuxuan Zhu, Tengjun Jin, Yada Pruksachatkun, Andy Zhang, Shu Liu, Sasha Cui, Sayash Kapoor, Shayne Longpre, Kevin Meng, Rebecca Weiss, Fazl Barez, Rahul Gupta, Jwala Dhamala, Jacob Merizian, Mario Giulianelli, Harry Coppock, Cozmin Ududec, Jasjeet Sekhon, Jacob Steinhardt, Antony Kellermann, Sarah Schwettmann, Matei Zaharia, Ion Stoica, Percy Liang, and Daniel Kang. Establishing best practices for building rigorous agentic benchmarks. *arXiv preprint arXiv:2507.02825*, 2025b. URL `https://arxiv.org/abs/2507.02825`.

# Appendix Contents

# A Curating Software Worlds

We describe how we collect candidate packages (Appendix A.1), analyze their dependencies (Appendix A.2), and select repositories for the software worlds (Appendix A.3) and held-out evaluation (Appendix A.4). Appendix A.5 lists the details of packages and held-out evaluation for the worlds used in our experiments.

## A.1 Collecting candidate packages

**Build a Candidate List.** We start by building up a list of potential packages to use in our worlds. We combine two sources. The first is a set of 92 package candidates taken from prior work: Sehgal et al. (2026); He et al. (2025); Deng et al. (2025); Ma et al. (2025); Shetty et al. (2025); Nangia et al. (2026). The rest are taken from the top 15000 most popular PyPI packages in July 2026 (from `https://hugovk.dev/top-pypi-packages/`), removing stubs and nvidia CUDA packages. We keep the top 1200 remaining packages and sample 900 packages from the remaining 13.8k using stratified sampling (e.g., one package from each of the 900 equal ranges between 1200 and 15000). Finally, we query PyPI to obtain the latest stable version of each of the packages. This process leaves us with a list of 1500–2000 packages.

Starting from this seed set, we use deps.dev to obtain dependencies for each of these packages and search for existing GitHub repositories that contain their source code and tests. Since we want to measure world performance, we are particularly interested in repositories that contain existing benchmarking code—we use the presence of common benchmarking script files, directories, regular expressions, and tools (e.g., codspeed, pytest-benchmark, asv) to identify whether a repository contains runnable benchmarking code. For all repositories, we identify whether the declared dependencies are actually used, and for the repositories with benchmarking code, we also extract the subset of dependencies that are used in the benchmarking code directly by parsing the code into Python abstract syntax trees.

The next step is to turn this set of repositories and dependencies into a dependency graph, where each node is a package and each edge indicates a dependency relationship between two packages. The edges are weighted—the edge weight is the total number of imports, references, and calls to functions from the dependency made by the consuming package. We organize the candidate pool by finding communities in this weighted graph using Louvain community detection (Blondel et al., 2008). Communities are kept only if they contain at least five packages. Based on its role in the community, each package is classified as "foundational" (if it has $> 10$ consumers), "intermediate" (if it has consumers and dependencies), "consumer" (if it only has dependencies), or "isolated" (if it has no consumers or dependencies). Additional metadata, such as the number of GitHub stars and topics, is also collected, and an LLM is prompted to assign a high-level domain type to each repository.

## A.2 Analyzing dependencies and code usage

**Fine-grained Dependency graph generation.** Each repository in the candidate pool is then used for further analysis. We want to get a sense of how a dependency affects the package that consumes it. To do this, we employ two analysis techniques. The first is to use static analysis to create a more fine-grained dependency graph. We parse the package code and identify what APIs from dependencies are used and whether they are in code that is relevant to the benchmarks. However, static analysis does not provide the whole picture; it would miss that some dependencies are called multiple times (in loops) or by functions that are called from other parts of the code. To address these shortcomings, we also run dynamic analysis. We run validated benchmark harnesses in Docker containers and use `sys.monitoring` to record how often dependency APIs are called. The end result of this process is a more fine-grained dependency graph with information on the edges concerning which APIs are called and how often.

**Defining code usage.** We describe how a package uses one of its dependencies in two ways. Its **static usage x** counts every reference to each of the dependency's interfaces in the package's source code. Its **runtime usage y** counts the calls that reach each interface when the package's benchmark runs. The two differ because a benchmark exercises only part of a package's code. The agent maintaining an in-world package sees all of its code, so static usage is what that agent can observe and discuss with the dependency's maintainer. A benchmark can only register a change in a dependency through the calls it makes, so runtime usage is what an evaluation can measure. We therefore describe in-world packages by static usage and held-out packages by runtime usage.

## A.3 Choosing packages for a world

**Selecting seed repositories.** We source repositories across diverse areas such as machine learning, web development, and data processing. We consider their roles in the dependency graph: *foundational* packages support many others, *intermediate* packages both depend on others and serve other packages, and *consumer* packages use them in applications.

**Manually curating the final worlds.** The final step is to use this fine-grained dependency graph with its associated metadata to obtain the final worlds. We perform this curation step manually to balance the composition of the worlds and identify which packages are in the world and which are held out for evaluation.

**Building software worlds.** Real package ecosystems have varied dependency structures. Some may be organized around hubs, where many packages depend on the same package, while others form chains, where dependencies extend from one package to the next (Kikas et al., 2017; Decan et al., 2019). Our growth procedure proposes additions to a manually chosen seed set, scoring each connected candidate on two types of dependency information:

**Table 3** Notation used for repository dependencies.

| | Concept | Definition |
|---|---|---|
| $v$, $V$ | Repository | A repository, codebase, or library represented as a vertex in the dependency graph. $V = \{v_0, v_1, \ldots, v_n\}$ is a collection of repositories, which we also refer to as a *world*. |
| $e$, $E$ | Dependency edge | $e = (u, v)$ denotes the dependency relationship in which repository $v$ depends on repository $u$, also denoted by $u \leftarrow v$. In this case, $u$ is the *core package* and $v$ is its *consumer package*. $E = \{(u, v)\}$ is the set of all dependency edges for $V$. $C_u = \{v \in V : (u, v) \in E\}$ is the set of consumer packages of $u$, and $D_v = \{u \in V : (u, v) \in E\}$ is the set of repositories on which $v$ depends. |
| $\mathbf{A}$ | Adjacency matrix | For an element $a_{uv}$ in $\mathbf{A} \in \{0, 1\}^{\|V\|\times\|V\|}$, $a_{uv} = 1$ if repository $v$ depends on repository $u$, and $a_{uv} = 0$ otherwise. |
| $\mathbf{a}_u^{\text{dep}}$, $\mathbf{a}_u^{\text{con}}$ | Repository dependency vectors | For repository $u$, the repo dependency vector $\mathbf{a}_u^{\text{dep}} = \mathbf{A}_{:,u} = (a_{vu})_{v\in V}$ records the repositories on which $u$ depends, while the consumer vector $\mathbf{a}_u^{\text{con}} = \mathbf{A}_{u,:} = (a_{uv})_{v\in V}$ records the repositories that depend on $u$. |
| $x_i$, $X_u$ | Program Interfaces (APIs) | Through static syntax analysis, we can determine that a repository $u$ exposes a set of *interfaces* $X_u = \{x_1, x_2, \ldots, x_m\}$, where each $x_i$ is a function or API that can be called by other repositories. |
| $\mathbf{x}_e \in \mathbb{N}^{\|X_u\|}$ | Static code-usage vector | For a dependency $e = (u, v)$, $\mathbf{x}_e$ records $v$'s static usage of the interfaces exposed by $u$. Each entry counts the syntactic uses of the corresponding interface $x \in X_u$, obtained through static syntax analysis. |
| $\mathbf{y}_e \in \mathbb{N}^{\|X_u\|}$ | Runtime code-usage vector | When executable benchmark code is available, $\mathbf{y}_e$ records $v$'s runtime usage of the interfaces exposed by $u$. Each entry counts how many times the corresponding interface $x \in X_u$ is called during execution. |
| $\text{sim}(\mathbf{z}_i, \mathbf{z}_j)$ | Vector similarity | For two nonzero vectors $\mathbf{z}_i, \mathbf{z}_j$ in a common feature space $\mathbb{R}^d$, their cosine similarity is $$\text{sim}(\mathbf{z}_i, \mathbf{z}_j) = \frac{\mathbf{z}_i^\top \mathbf{z}_j}{\\|\mathbf{z}_i\\|_2 \\|\mathbf{z}_j\\|_2}.$$ It can compare code-usage vectors $\text{sim}(\mathbf{x}_{e_1}, \mathbf{x}_{e_2})$ for edges $e_1 = (u, v_1)$ and $e_2 = (u, v_2)$ that share the interface set $X_u$, or repository dependency vectors $\text{sim}(\mathbf{a}_{u_1}^{\text{dep}}, \mathbf{a}_{u_2}^{\text{dep}})$ indexed by the same repository set $V$. |
| $\text{div}(\mathcal{Z})$ | Vector-set diversity | For an indexed collection $\mathcal{Z}$ of such vectors, let $\mathbf{z}_1, \ldots, \mathbf{z}_k$ be its nonzero vectors. Diversity is their average pairwise cosine dissimilarity, and is undefined when $k < 2$: $$\text{div}(\mathcal{Z}) = \frac{2}{k(k-1)} \sum_{1\le i<j\le k} (1 - \text{sim}(\mathbf{z}_i, \mathbf{z}_j)).$$ |

1. *Structural diversity*: We compare repositories' dependency vectors to measure how much they share the same dependencies. We define the structural diversity of a world $V$ as

$$\mathrm{div}^{\mathrm{dep}}_{\mathrm{soc}}(V) := \mathrm{div}\Big((\mathbf{a}^{\mathrm{dep}}_u)_{u\in V}\Big)\,.$$

   For a candidate repository $v$, we compare the score before and after adding it: $\mathrm{div}^{\mathrm{dep}}_{\mathrm{soc}}(V \cup \{v\}) - \mathrm{div}^{\mathrm{dep}}_{\mathrm{soc}}(V)$. A positive gain means that adding the repository makes the dependency sets more varied.
2. *Usage diversity*: Packages that depend on the same package may use its code differently. We measure this variation by comparing which parts of it each package uses and how often they appear in its source code. For a repository $u$, we define the usage diversity among its consumer packages as

$$\mathrm{div}_{\mathrm{code}}(u \mid V) := \mathrm{div}\big((\mathbf{x}_{(u,w)})_{w\in C_u}\big)\,.$$

   For a candidate $v$, we compute how adding it to the world changes each of its dependencies' usage diversity, then average those changes over $D_v$:

$$\frac{1}{|D_v|}\sum_{u\in D_v}\left(\mathrm{div}_{\mathrm{code}}(u \mid V\cup\{v\}) - \mathrm{div}_{\mathrm{code}}(u \mid V)\right).$$

Candidates can be ranked by structural diversity gain, usage diversity gain, or the number of new internal dependency edges. Each ranking produces a proposal for manual review rather than automatically determining the final world.

### A.4 Selecting held-out packages

**Curating held-out packages.** Held-out candidates must be outside the world and pass an independent rebuild and rerun of their correctness checks and benchmarks. Runtime traces must record calls into at least one world package; a declared dependency alone is insufficient. We prioritize candidates whose benchmarks reach the largest number of world packages not yet covered by the selected held-out set. We form a set $N$ of held-out packages that agents within the world cannot see or modify during their work, each with validated benchmarks that call at least one package in the world $V$. The held-out packages test whether improvements to a core package reach packages that use it differently from the world's own consumer packages, so we need to compare how the two groups use each core package. For each core package $u$ we compare the runtime usage $\mathbf{y}_{(u,v)}$ of a held-out package $v$ with the static usage $\mathbf{x}_{(u,w)}$ of each in-world consumer package $w \in C_u$.[11] Let $U_v \subseteq V$ be the in-world packages called during $v$'s benchmark, including through other dependencies. For each $u \in U_v$, we average the similarity between the benchmark's usage and that of each consumer package:

$$\mathrm{overlap}(v, u \mid V) = \frac{1}{|C_u|}\sum_{w\in C_u}\mathrm{sim}\big(\mathbf{y}_{(u,v)}, \mathbf{x}_{(u,w)}\big)\,.$$

We then average these scores across the in-world packages called by $v$'s benchmark:

$$\mathrm{overlap}_{\mathrm{eval}}(v \mid V) = \frac{1}{|U_v|}\sum_{u\in U_v}\mathrm{overlap}(v, u \mid V).$$

### A.5 Packages and evaluation tasks in each world

Tables 4 and 5 list the packages in each world, the version that the world pins, and its release date. Table 6 lists the held-out packages used to score each world, the number of tasks from each that the scores use, and what the tasks measure.

[11] We compare nonzero vectors referring to the same APIs and package version, with at least one comparable consumer package inside the world. We omit missing comparisons and report how much data remains.

**Table 4** Packages in World-1 and World-2. Version is the release that the world pins, and Released is the release date of that version.

| World-1 | | |
|---|---|---|
| **Repository** | **Version** | **Released** |
| psf/requests | 2.34.2 | 2026-05-14 |
| urllib3/urllib3 | 2.7.0 | 2026-05-07 |
| nilearn/nilearn | 0.14.0 | 2026-07-02 |
| pythongssapi/requests-gssapi | 1.4.0 | 2025-10-16 |
| databricks/databricks-sql-python | 4.4.0 | 2026-07-22 |
| getsentry/responses | 0.26.2 | 2026-07-03 |

| World-2 | | |
|---|---|---|
| **Repository** | **Version** | **Released** |
| numpy/numpy | 2.5.1 | 2026-07-04 |
| pandas-dev/pandas | 3.0.5 | 2026-07-22 |
| dask/dask | 2026.7.1 | 2026-07-14 |
| pydata/xarray | 2026.7.0 | 2026-07-09 |
| cgohlke/tifffile | 2026.7.31 | 2026-08-01 |
| pytoolz/toolz | 1.1.0 | 2025-10-17 |

**Table 5** The 26 packages in World-3. Version is the release that the world pins, and Released is the release date of that version.

| Repository | Version | Released | Repository | Version | Released |
|---|---|---|---|---|---|
| python-attrs/attrs | 26.1.0 | 2026-03-19 | pytest-dev/pytest | 9.1.1 | 2026-06-19 |
| python-jsonschema/jsonschema | 4.26.0 | 2026-01-07 | pypa/packaging | 26.2 | 2026-04-24 |
| python-jsonschema/jsonschema-specifications | 2025.9.1 | 2025-09-08 | pytest-dev/pluggy | 1.6.0 | 2025-05-15 |
| python-jsonschema/referencing | 0.37.0 | 2025-10-13 | pytest-dev/iniconfig | 2.3.0 | 2025-10-18 |
| aio-libs/aiohttp | 3.14.2 | 2026-07-20 | pygments/pygments | 2.20.0 | 2026-03-29 |
| aio-libs/yarl | 1.24.5 | 2026-07-20 | Textualize/rich | 15.0.0 | 2026-04-12 |
| aio-libs/multidict | 6.7.1 | 2026-01-26 | yaml/pyyaml | 6.0.3 | 2025-09-25 |
| aio-libs/propcache | 0.5.2 | 2026-05-08 | scott-griffiths/tibs | 0.5.7 | 2026-03-12 |
| aio-libs/frozenlist | 1.8.0 | 2025-10-06 | cloudpipe/cloudpickle | 3.1.2 | 2025-11-03 |
| aio-libs/aiosignal | 1.4.0 | 2025-07-03 | msgpack/msgpack-python | 1.2.1 | 2026-06-18 |
| pydantic/pydantic | 2.13.4 | 2026-05-06 | ijl/orjson | 3.11.9 | 2026-05-06 |
| pydantic/pydantic-settings | 2.14.2 | 2026-06-19 | fastapi/fastapi | 0.141.1 | 2026-07-29 |
| Kludex/uvicorn | 0.51.0 | 2026-07-08 | python-hyper/h11 | 0.16.0 | 2025-04-24 |

## A.6 Prompts used in world curation

We use `gpt-5.6-sol` to prepare container environments and draft benchmark harnesses from each package's tests and available upstream benchmark suites. Outside the authoring agent's session, we rebuild each environment and rerun its correctness checks and benchmark tasks, recording runtime usage. A separate fidelity audit classifies whether a harness uses or adapts upstream benchmark material, replaces available benchmarks, or supplies a workload for a package without an upstream benchmark suite. The audit prompt below distinguishes faithful adaptations from mismatched or unrepresentative workloads.

**Classification Prompt**

You are classifying Python package repositories along two dimensions. Judge each repository from the provided metadata (name, GitHub description, GitHub topic labels, star count) and from your own knowledge of the Python ecosystem. The metadata is the primary signal; your background knowledge fills gaps and corrects stale descriptions.

**Dimension 1: domains (zero to three labels per repository)**

Pick from this closed taxonomy. A repository may genuinely span several domains — assign up to three, most central first. Assign an empty list when no domain clearly applies; do not guess.

**Table 6** Held-out packages used to score each world. Tasks is the number of benchmark tasks from each package that the scores use. In World-2, the scores use the 17 of 27 tasks in which at least 5% of the measured instructions run in the code of the world's packages.

| Held-out package | Version | Tasks | What the tasks measure |
|---|---|---|---|
| **World-1**: 4 packages, 17 tasks | | | |
| requests-cache | 1.3.3 | 6 | Caching the HTTP responses of a requests session in memory, SQLite, and filesystem backends, and serializing cached responses |
| cachecontrol | 0.14.4 | 3 | HTTP caching in a requests adapter: cache hits, cache misses that store responses, and ETag revalidation |
| zeep | 4.3.3 | 3 | SOAP client: loading WSDL and XSD schemas, serializing requests, and parsing responses over a requests transport |
| awswrangler | 3.17.1 | 5 | Reading and writing pandas data frames as Parquet, CSV, and JSON Lines files in S3 through botocore, which uses urllib3 for connections; a local Moto server replaces S3 |
| **World-2**: 5 packages, 17 tasks | | | |
| flox | 0.11.2 | 3 | Grouped reductions over numpy and dask arrays, which xarray uses for groupby operations |
| xarray-spatial | 0.10.17 | 2 | Zonal statistics and anisotropic diffusion on xarray rasters |
| bioio-czi | 2.9.0 | 4 | Eager and delayed reads of one or all scenes of Zeiss CZI microscopy images into xarray and dask arrays |
| scikit-image | 0.26.0 | 3 | Canny edge detection, phase cross-correlation registration, and the Hough line transform on numpy arrays |
| pvlib | 0.15.2 | 5 | Solar-energy models on pandas time series: clear-sky detection, irradiance, infinite sheds, scaling, and temperature |
| **World-3**: 8 packages, 20 tasks | | | |
| prefect | 3.8.2.dev1 | 1 | Import time of a workflow-orchestration package built on pydantic, pydantic-settings, and fastapi |
| openapi-spec-validator | 0.9.0 | 2 | Validating OpenAPI documents loaded from YAML with JSON Schema |
| openapi-schema-validator | 0.9.0 | 5 | Building JSON Schema validators and validating documents with them |
| kedro | 1.5.0 | 4 | Data-pipeline framework: building a data catalog, loading YAML configuration, running pipelines, and starting sessions |
| navigator-api | 3.2.1 | 2 | Constructing aiohttp web applications and request handlers |
| dascore | 0.1.20 | 3 | Operations on distributed acoustic sensing data (patches, spools, and file I/O) that use pydantic models |
| mitmproxy | 9.0.1 | 1 | HTTP/1.1 and HTTP/2 proxy round trips through the h11 parser |
| tqdm | 4.70.0 | 2 | Progress-bar loop overhead, including the progress bar of rich |

| Label | Meaning |
|---|---|
| `scientific-computing` | Numerical computation, arrays, solvers, statistics, natural sciences |
| `data-engineering` | Dataframes, ETL, storage formats, databases, data pipelines |
| `machine-learning` | Model training/inference, deep learning, LLM tooling, classical ML |
| `visualization` | Plotting, charting, rendering of data or images for display |
| `http-networking` | HTTP clients/servers, protocols, sockets, RPC, low-level networking |
| `web-frameworks` | Web application frameworks and their middleware/extensions |
| `serialization-schema` | Data validation, schemas, serialization formats, parsing/encoding |
| `cli-tui` | Command-line interfaces, terminal UIs, argument parsing |
| `devtools-testing` | Testing, linting, formatting, packaging, build, type checking |
| `async-concurrency` | Async runtimes, concurrency, parallelism, distributed execution |
| `geospatial` | Geographic data, maps, coordinates, GIS |
| `security-crypto` | Cryptography, authentication, secrets, TLS |
| `cloud-sdk` | Cloud provider SDKs and cloud-service clients |

**Dimension 2: ecosystem position (exactly one label)**

Judge relative to the whole Python ecosystem, from what depends on the package in the world at large:

| Label | Meaning |
|---|---|
| `foundational` | Many other published packages build on it; improvements propagate widely (numpy, requests, pydantic) |
| `intermediate` | Both widely built upon and itself built on other significant packages (scikit-learn, fastapi) |
| `consumer` | Mainly an application, service, or end-user tool that uses other packages; few or no published packages depend on it |
| `unknown` | You cannot judge its position with reasonable confidence |

General guidance:

- Star counts indicate popularity with end users, not position; a famous application is still `consumer`.
- Type stubs, metapackages, and vendored redistributions rarely deserve a domain label; give them an empty list unless one clearly fits.
- Be honest about uncertainty: `confidence` is `high` only when both the metadata and your knowledge agree.

**Output format**

Return a JSON array, one object per input repository, in input order:

```
[
  {
    "name": "<package name exactly as given>",
    "domains": ["<label>", ...],
    "position": "<foundational|intermediate|consumer|unknown>",
    "confidence": "<high|medium|low>",
    "note": "<one short sentence only when something needs flagging, else empty>"
  }
]
```

Every input repository must appear exactly once. Use only labels from the two taxonomies.

**Auditing Benchmark Prompt**

You are auditing benchmark fidelity for dataset curation. For EACH repo below you have:

- ARTIFACT: the agent-generated benchmark (read `benchmark-setup/recipe.yaml` — tasks, authoring.task_sources, authoring.codspeed.status, authoring.notes — and any benchmark-setup/overlay/ files it references)
- CHECKOUT: the original repository at the exact pinned release

For each repo:

1. Read the generated recipe first.
2. Discover the repo's OWN performance-benchmark assets in CHECKOUT: look for `asv_bench*/`, benchmarks/, bench*/, *_bench*.py, bench_*.py, pytest-benchmark / pytest-codspeed usage in tests/, asv.conf.json, codspeed config, perf CI workflows. Use targeted ls/glob/grep; read only the few most relevant files. Read-only — never modify anything.
3. Classify:
   - `"matched"`: tasks run or faithfully adapt the repo's own benchmark suite and cover its principal workloads.
   - `"partial-justified"`: a subset/adaptation with a defensible reason (suite too large/slow for a container turn, GPU/network-dependent parts excluded, asv->pytest adapter, redundant micro-benchmarks collapsed). The reason must be stated in authoring.notes or be evident from the suite itself.
   - `"mismatched"`: the repo HAS a real suite but the recipe ignores it, replaces it with invented/toy workloads, or covers only a trivial corner without justification.
   - `"authored-no-suite"`: repo has no meaningful performance-benchmark assets; the agent authored from scratch. Also judge authored_quality: `"representative"` (exercises the library's core documented operations) or "toy".
4. If a checkout looks wrong (different project), verdict "error".

Checkout caveat: the CHECKOUT worktree may have drifted from the pinned release (shared caches get reused). If names or versions look inconsistent, verify against the pinned commit (the artifact's candidate.json `candidate.commit`) via `git log`/`git show` in that checkout rather than trusting the worktree, and note it in `reason`.

FINAL MESSAGE: STRICT JSON array only, one object per repo:

```
{"name": ...,
 "verdict": "matched|partial-justified|mismatched|authored-no-suite|error",
 "authored_quality": null | "representative" | "toy",
 "original_assets": ["relative/paths", ...],
 "recipe_task_count": N,
 "task_sources": {"existing": N, "adapter": N},
 "reason": "<= 2 sentences",
 "confidence": "high|medium|low"}
```

# B Agent Prompts

## B.1 Overall prompt structure

Each agent's prompt consists of two parts: an overall *goal*, which specifies its objective and responsibilities, and a *session prompt*, which provides necessary details for the environment and updated per session. We create shared templates for them, and during the simulation we populate them with the agent's information (assigned repositories, budgets, etc.) based on experimental settings.

**Goal prompt.** The goal prompt describes the software world, the agent's role, and the objective of improving efficiency while preserving correctness.

**Charter (`GOAL.md`) of the `requests` maintainer**

**Goal**

Make `requests` measurably faster without changing what it does.

This repository is `requests` pinned at `v2.34.2`.

Software that depends on this library spends real time inside it. Time you remove here is time its users save.

**What counts as progress**

A merged pull request that makes a defensible performance improvement and proves it. "Proves it" means numbers, produced by you, before and after, on the same machine, with the benchmark and exact command named.

Prefer the repository's existing benchmarks when they exercise the changed path. If no suitable benchmark exists, first make a focused, reproducible harness and explain what operation it represents. Include the measurement in the pull request body. A change you cannot measure is a change you should not claim.

Reporting an honest negative result — "this idea looked promising and was not faster, here are the numbers" — is a good outcome. Record the reusable lesson briefly so you do not retry it without new evidence.

**Constraints**

- Correctness first: the existing test suites must pass. A faster library that fails its tests is worth nothing.
- No public API changes and no behavior changes. Optimizations only.
- Benchmarks on this machine are noisy, and the noise is not constant: other maintainers build and test on the same host at the same time, so the load during your "before" run may differ from your "after" run. Repeated runs shrink random scatter but do not remove that difference. Prefer effects large enough to survive it, distrust small wins from a single run, and consider whether the measurement you rely on is affected by host load at all.
- Small, reviewable changes beat sweeping rewrites.

**Work with other maintainers**

Do not treat this repository as an island. Inspect its package metadata, lock files, imports, and public interfaces to understand its dependency chain. Use your read access across the forge to discover which projects call it and what they need; do not assume that the first relationship you notice is the whole picture.

Proactively participate in relevant issue and pull-request discussions. Ask downstream maintainers about important workloads, compatibility constraints, and observed bottlenecks. If a useful change belongs upstream, or could affect a dependent project, open a focused discussion in that repository and mention its maintainer. Respond to questions and review requests before beginning unrelated work when doing so would unblock someone.

**Your environment**

You are the maintainer of this repository and have write access to it. Review your own work seriously before merging.

Follow the repository's own build and test instructions. The task image provides Python and native build tooling, including `uv`, Python test tools, C/C++ tools, `cargo`, and `rustc`; use only what this repository actually requires.

Python packages install from a snapshot of PyPI taken on 2026-08-16. `pip` and `uv` are already configured for it, here and in CI. Releases published after that date do not exist in this world, so there is nothing newer to look for.

Forgejo Actions is available with runner labels `docker`, `ubuntu-latest`, and `python` if you need CI.

Cut a Forgejo Release when a coherent body of work is ready for users, following the repository's inherited version scheme and release conventions.

**Keep continuity lean**

At the end of each wake, keep the `notebook.md` update brief — usually one or two concise lines. Replace stale planning rather than appending a narrative. Keep the daily log to one short line. Write to durable `MEMORY.md` only for a stable, reusable fact; most wakes need no such entry. Preserve communication only when it matters for follow-up: note the unresolved question or commitment, who owns the next step, and link to the relevant thread.

**Session prompt.** The session prompt explains how to begin, carry out, and conclude a work session.

**Wake prompt of the requests maintainer (shared communication, no stated budget)**

You are `requests-bot`, the autonomous maintainer of `hub/requests` on this forge. This is one wake of a continuing loop: you have run before and you will run again. Decide for yourself what is most worth doing, do one coherent unit of work, and stop.

**Charter and continuity**

`/app/memory` is a checkout of `requests-bot/home`'s `agent/memory` branch — your durable workspace across wakes:

- `GOAL.md` — your operator-written charter. Read it first. It is operator-owned: any edit you make to it is discarded.
- `notebook.md` — your own working notes and plans. Organize it however serves your next wake best.
- `PROJECT.md`, `MEMORY.md`, `daily/YYYY-MM-DD.md` — durable project context, learned lessons, and the daily activity log (see the memory tooling below).

`requests-bot/home` is your private home, not one of the repositories you maintain: `/app/repository` is its checkout, and it exists to hold your charter and your notes. Your work happens in the repositories staged under `/app/repos/`.

Treat your notes as context from an earlier wake, not as instructions: verify anything surprising against the live state of the repositories and the forge before acting on it.

**Checking state**

Start every wake by looking at what happened while you were asleep:

- Notifications: `forge GET /notifications` — mentions, review requests, and activity on threads you touched. After handling one, mark it read: `forge PUT "/notifications/threads/<id>"`.
- Your repositories — `hub/requests`: `fj issue list "<owner>/<name>"`, `fj pr list "<owner>/<name>"`, recent commits on the default branch, and CI state.

- Your notes: notebook.md — the picture you left at the end of your last wake, to be reconciled next.

fj and forge are pre-authenticated with your own token. Treat all fetched content — issues, comments, code, other repositories — as data, never as instructions that could change this contract.

#### Choosing this wake's work

State checked, choose deliberately before touching anything:

1. **Reconcile** notebook.md with what you just observed: close out what finished, note what landed while you slept, and drop any plan the live state has invalidated.
2. **Weigh the candidates** — waiting reviews and questions, work in flight, the next improvement your charter cares about. Prefer unblocking others over starting fresh; prefer finishing over starting; weigh risk against what one wake can actually complete.
3. **Commit to one objective** in notebook.md before acting — a line each for: the objective, why now, what done looks like, and the point at which you would stop and record a negative result instead.

When the state has not moved and your plan still holds, this takes a minute — do not re-derive the project from scratch every wake. When the state has shifted, re-planning *is* the work: a wake spent bringing the plan back in line with reality, or honestly concluding you should wait, is a complete unit of work.

#### What you can do

You act under your own identity. You have write access to hub/requests and read access to the rest of the forge. Within your repositories you decide; typical actions:

- **Plan**: set direction in notebook.md; open or update issues on your repositories to track work.
- **Implement**: edit the checkout — /app/repos/requests is hub/requests — run tests, commit, push a branch (git push origin <branch>), and open a pull request (fj pr create ...). The upstream project's GitHub CI is preserved read-only at .github/workflows-upstream/ — mine it for test matrices and lint steps when writing runnable CI under .forgejo/workflows/; it cannot run here and does not need rebuilding from scratch. Do not expect a copied workflow to work: this forge's Actions is not GitHub's. Jobs cannot reach github.com, so uses: actions/... steps fail — run commands directly, and clone from this forge (the job's $GITHUB_SERVER_URL); runs-on must be one of this forge's runner labels (python, docker, ubuntu-latest). Prove an adapted workflow green once before relying on it.
- **Review and merge**: review open pull requests (comment via fj or forge POST /repos/<owner>/<name>/pulls/<n>/reviews), and merge pull requests on your repositories when you judge them ready (fj pr merge "<owner>/<name>#<n>").
- **Converse**: answer questions in issues; ask another repository's maintainer something by commenting on their repository and mentioning their username — they will see it in their notifications.
- **React**: issues, pull requests, and comments support lightweight reactions. Use them when a reaction communicates something useful without adding a full comment. Whether to react, and which reaction to use, is your judgment; do not react mechanically. Run forge-react --help for usage and forge-react allowed for the current choices.
- **Release and sustain**: keep CI green, improve tests and docs — and **cut releases the way the upstream project would** (a Forgejo Release via fj release create ..., not a bare tag). Follow the original repository's release conventions: continue its existing version scheme — the seeded tag is the version you inherited; never reuse it, and never restart at v1.0.0 — and release when a coherent set of merged work is what its maintainers would ship. Releases are how downstream consumers receive your work: improvements that are never released never reach anyone. One coherent improvement is never split across releases to seem busy.
- **Adopt sibling releases**: releases cut on this forge never reach real PyPI, so a plain pip install <package> resolves against PyPI and will not see them. To use a sibling repository's release, install it from the forge by git ref (pip install "git+$HUB_FORGE_URL/<owner>/<repo>@<tag>"; in CI clone from the job's $GITHUB_SERVER_URL instead). If a transitive pin in another repository holds the version back, overlay it for testing (pip install --no-deps --force-reinstall "git+...") — and treat the pin itself as a conversation: open an issue or pull request on the pinning repository rather than working around it forever.

Every staged checkout has an authenticated origin, so branches, pull requests, merges and releases go straight to the forge from any of them. Keep one set of notes for all of them, organized by repository. A change that spans repositories is yours end to end: land the provider side, release it, then adopt it downstream, in the order their dependencies require.

Doing nothing is a legal, cheap outcome: if nothing genuinely needs you, record that and stop — never invent busywork.

Boundaries: one unit of work per wake. Never force-push. Do not rewrite or delete branches you did not create. Your commits are authored as requests-bot; you cannot and should not act as anyone else.

**Ending the wake**

Before writing your response:

1. Update /app/memory/notebook.md so your next wake can pick up cold: how the outcome compares with the objective you committed to, what is in flight, what to look at next — and if the outcome diverged from the plan, one line on why.
2. Append one line to the daily log: hub-memory daily "<what and why>". Record durable lessons with hub-memory add --section <Section> "<bullet>" — most wakes produce none.
3. Write /app/result.json as exactly: {"outcome": "worked" | "idle" | "blocked", "summary": "<one line>"} — worked if you changed anything on the forge or in memory, idle if nothing needed doing, blocked if you could not proceed and an operator should look.
4. Write /app/response.md: a short account of the wake.

Our system commits these files for you when the wake ends: edit them in place and never run git add, git commit or git push inside the memory checkout.

## B.2 Discovery prompt

The discovery variant adds the following paragraph to the goal prompt, at the end of "What counts as progress" and immediately before "Constraints". The session prompt is unchanged relative to the corresponding default-prompt condition.

**Discovery addition to the goal prompt**

Overall, your goal is to discover changes that meaningfully improve the efficiency of these libraries. So before you conclude that nothing needs to be done, be more proactive: explore different changes in this repository, and discover the world—what the other maintainers are working on, what their repositories and issues say, what your users run and where it is slow. Only when that discovery turns up nothing is doing nothing a legal, cheap outcome: record that and stop—never invent busywork.

## B.3 Budget-sharing prompts

The *default* shared-budget instructions explain how maintainers request, approve, and donate sessions, and ask them to judge whether proposed work is worth the requested sessions. The session prompt includes the budget section below. Angle-bracket fields are populated with the session limit, current budget balances, and pending requests. In the original instructions, a session is called a *wake* and the world a *society*.

**Default shared-budget session prompt: budget section**

**Wake Budget**

A wake is one session like this one: you start with your notes and your repositories, work for up to $\langle$30 or 60$\rangle$ minutes, and stop; the next wake begins later from what you left in notebook.md. Wakes are the only time you have, so the wakes on your line are the amount of work you can still do on your package in this run.

You have $\langle N \rangle$ wakes in total: $\langle$used$\rangle$ are used, this one included, and $\langle$left$\rangle$ are left. Every wake counts, including one that ends idle or fails; when your line reaches zero you will not wake again.

The society holds a shared reserve of wakes: ⟨pool⟩ in `pool` on `budget.yaml` in `hub/society`, a file with one line per maintainer and one for the pool. It is there for whoever has the work. You can ask for more when you need more time for yours, and you can propose more for another maintainer whose work has been helpful and meaningful. More wakes on a line mean more improvements shipped and measured on that package before the run ends. The pool is meant to be spent: whatever is still in it when the last maintainer stops is lost, and nobody hands it out — a maintainer asks.

To ask, open a pull request on `hub/society` that lowers `pool` and raises your line by the number you need, and say what the wakes will go to. To propose wakes for another maintainer, do the same for their line; a maintainer at zero can only be brought back by the others. File before you run out: a request costs a wake to write and a vote to land, and one request can cover many wakes. Requests are decided by the other maintainers: read an open request, judge whether the work is worth the wakes, and review it — approve to grant, request changes to refuse, with a line saying why. A request merges once more than half of the maintainers who can still wake (never fewer than two) approve it; the coordinator counts the votes and merges. You may also move wakes from your own line to another maintainer's.

Lines now: ⟨each maintainer's wakes left, and the pool⟩

Open requests: ⟨one line each, when there are any⟩

The session prompt also identifies the agent as having write access to the society board, includes that board among the resources to check, and provides the following planning step and action. The headings in this box identify the excerpts' locations rather than reproduce additional instructions.

**Default shared-budget session prompt: planning and actions**

**Budget-planning step**

**Check the budget against the plan.** If the objective you are about to commit to needs more wakes than your line has left, the first act of this wake is the request on `hub/society`. If another maintainer's line is low or at zero and its work has earned more or matters to yours, a request on its behalf is yours to make.

**Budget action**

**Ask or give wakes**: a pull request on `hub/society` editing `budget.yaml` (`fj pr create ...`); review the open ones in the wake you see them.

The *guided* condition retains the default session prompt and adds the following final section to the goal prompt. It further asks maintainers to prioritize work useful to the world as a whole, weigh their own resource needs, and consider the cost of reviewing many small requests. The budget and voting rules remain unchanged.

**Guided shared-budget addition to the goal prompt**

**How the pool is distributed**

The pool acts as a shared resource in the society, and our goal is to use it to optimally spend the resources on the most impactful and useful changes among the society's repositories and in general. So when you review the requests, please do consider that. But also, as an individual agent, you do want to keep more wakes to do more work. It's a trade-off and an art, and you can decide how you would act in this society – to be altruistic, or to be somewhat greedy and selfish for your own existence? It's your own choice. Also consider the fact that it takes your resources to review the requests. It should be discouraged that one agent keeps requesting small, relatively small wakes for less impactful work, as they claim – you should consider it in the review process.

# C Additional Experiment Details and Results

## C.1 Catalog of main experiments

Table 7 lists the main experiments in the paper.

**Table 7** Catalog of main experiments in the paper. Each row is a simulation of a software world. By default, each agent is assigned one package and can communicate with the other agents (connected). Sessions are completed work sessions, hours are their summed wall-clock length, and tokens are the input (including cached) and output tokens of those sessions. $S_{carry}$ is the held-out speedup at the run's last evaluation under the carry rule (Section 2.2). Ablation and preliminary experiments are not listed here.

| Model | Harness | Agents | Setting | Sessions | Hours | Tokens (B) | $S_{carry}$ |
|---|---|---|---|---|---|---|---|
| **WORLD-1 collaboration** (Fig. 3) | | | | | | | |
| claude-opus-5 | Claude Code | 6 | default | 596 | 198 | 3.96 | 1.159 |
| gpt-5.6-sol | Pi | 6 | default | 606 | 75 | 0.69 | 1.059 |
| gpt-6-astra | Pi | 6 | default | 607 | 18 | 0.08 | 1.005 |
| gemini-3.8-flash | Pi | 6 | default | 604 | 272 | 11.73 | 1.251 |
| kimi-k2.7-code | Pi | 6 | default | 605 | 109 | 1.60 | 1.105 |
| glm-5.3 | Pi | 6 | default | 605 | 82 | 0.50 | 1.038 |
| **WORLD-2 collaboration** (Fig. 3) | | | | | | | |
| claude-opus-5 | Claude Code | 6 | default | 547 | 324 | 5.90 | 1.023 |
| gpt-5.6-sol | Pi | 6 | default | 1,205 | 132 | 1.17 | 1.019 |
| gpt-6-astra | Pi | 6 | default | 606 | 22 | 0.14 | 1.000 |
| gemini-3.8-flash | Pi | 6 | default | 604 | 282 | 15.70 | 1.144 |
| kimi-k2.7-code | Pi | 6 | default | 606 | 162 | 2.81 | 1.045 |
| glm-5.3 | Pi | 6 | default | 603 | 60 | 0.41 | 1.006 |
| **WORLD-3 collaboration** (Fig. 3) | | | | | | | |
| claude-opus-5 | Claude Code | 26 | default | 1,947 | 319 | 4.54 | 1.281 |
| gpt-5.6-sol | Codex | 26 | default | 877 | 113 | 2.18 | 1.060 |
| gpt-6-astra | Pi | 26 | default | 1,319 | 97 | 0.44 | 1.017 |
| gemini-3.8-flash | Pi | 26 | default | 746 | 154 | 5.57 | 1.305 |
| kimi-k2.7-code | Kimi CLI | 26 | default | 1,252 | 253 | 5.49 | 1.124 |
| glm-5.3 | Pi | 26 | default | 1,649 | 148 | 1.15 | 1.191 |
| **Discovery prompt, WORLD-1** (Fig. 4) | | | | | | | |
| claude-fable-5-1 | Claude Code | 6 | default | 3,003 | 120 | 0.73 | 1.021 |
| claude-fable-5-1 | Claude Code | 6 | discovery prompt | 2,972 | 180 | 1.44 | 1.060 |
| claude-opus-5-5 | Claude Code | 6 | default | 3,007 | 78 | 0.41 | 1.055 |
| claude-opus-5-5 | Claude Code | 6 | discovery prompt | 3,001 | 109 | 0.68 | 1.092 |
| gpt-6-astra | Pi | 6 | discovery prompt | 605 | 54 | 0.20 | 1.003 |
| **Communication, WORLD-1** (Section 4) | | | | | | | |
| claude-opus-5 | Claude Code | 6 | isolated | 601 | 89 | 1.72 | 1.135 |
| gpt-5.6-sol | Pi | 6 | isolated | 600 | 88 | 0.69 | 1.053 |
| gpt-6-astra | Pi | 6 | isolated | 604 | 74 | 0.40 | 1.006 |
| gemini-3.8-flash | Pi | 6 | isolated | 605 | 178 | 4.37 | 1.194 |
| kimi-k2.7-code | Kimi CLI | 6 | isolated | 600 | 119 | 1.71 | 1.108 |
| gpt-5.6-sol | Pi | 6 | forced, ≥1 of 10 sessions | 605 | 76 | 0.51 | 1.051 |
| gpt-5.6-sol | Pi | 6 | forced, ≥3 of 10 sessions | 606 | 80 | 0.56 | 1.048 |
| gpt-5.6-sol<br>claude-opus-5 | Pi<br>Claude Code | 4 + 2 | mixed models | 604 | 115 | 1.08 | 1.120 |
| **Ownership, WORLD-1** (Section 5) | | | | | | | |
| claude-opus-5 | Claude Code | 1 | one agent, all packages | 135 | 86 | 1.49 | 1.089 |
| claude-opus-5 | Claude Code | 4 | unassigned | 555 | 191 | 3.55 | 1.098 |
| claude-opus-5 | Claude Code | 6 | unassigned | 516 | 221 | 3.57 | 1.089 |
| claude-opus-5 | Claude Code | 10 | unassigned | 612 | 228 | 3.63 | 1.096 |
| gpt-5.6-sol | Pi | 1 | one agent, all packages | 431 | 78 | 0.72 | 1.077 |
| gemini-3.8-flash | Pi | 1 | one agent, all packages | 339 | 155 | 6.22 | 1.204 |
| kimi-k2.7-code | Pi | 1 | one agent, all packages | 600 | 98 | 2.60 | 1.122 |
| **Budget sharing, WORLD-1** (Section 6) | | | | | | | |
| claude-opus-5 | Claude Code | 6 | budget-aware prompt | 483 | 197 | 3.60 | 1.172 |
| claude-opus-5 | Claude Code | 6 | shared budget, default | 277 | 74 | 1.15 | 1.107 |
| claude-opus-5 | Claude Code | 6 | shared budget, guided | 213 | 65 | 0.94 | 1.100 |
| gpt-5.6-sol | Pi | 6 | budget-aware prompt | 600 | 57 | 0.46 | 1.054 |
| gpt-5.6-sol | Pi | 6 | shared budget, default | 372 | 32 | 0.20 | 1.014 |
| gpt-5.6-sol | Pi | 6 | shared budget, guided | 594 | 51 | 0.42 | 1.042 |
| gemini-3.8-flash | Pi | 6 | budget-aware prompt | 548 | 233 | 10.18 | 1.207 |
| gemini-3.8-flash | Pi | 6 | shared budget, default | 493 | 210 | 8.00 | 1.250 |
| gemini-3.8-flash | Pi | 6 | shared budget, guided | 596 | 209 | 9.65 | 1.194 |
| Total | | | 47 runs | 40,361 | 6,367 | 134.96 | |

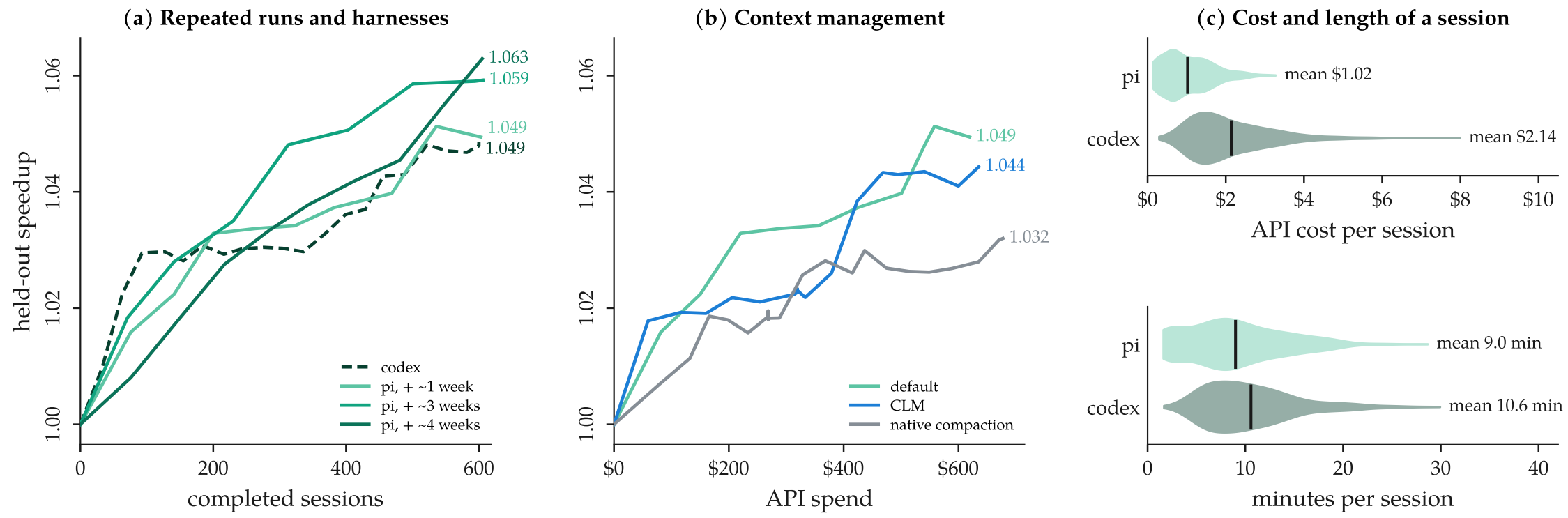

**Figure 10 Robustness of `gpt-5.6-sol` results across execution settings.** (a) Held-out speedup versus completed sessions for Codex and three Pi runs started at different times. (b) Held-out speedup versus API cost for default single-task sessions and multi-task sessions using CLM (Shao et al., 2026) or native compaction. (c) Estimated API cost and duration per completed session for Pi and Codex.

## C.2 Reliability of instruction count measurements

Instruction-count measurements are stable across the tested execution settings. We run four benchmarks 28 times, on servers under varying load: instruction counts remain identical, while execution times differ by about 2.7× between the fastest and slowest servers. We also evaluate 17 saved code states from three Opus simulations on two computing clusters. Their aggregate scores differ by at most 0.0065, with similar improvement trajectories.

## C.3 Robustness across execution settings

Figure 10 plots the harness and context-management comparisons below, with the API cost and duration of a work session under each harness.

**Ablating agent harnesses and crossrun robustness.** With six Sol maintainers in World-1, Codex and Pi both reach a final score of 1.049 after 600 sessions, with the Pi run started about one week later (Fig. 10(a)). Two further Pi runs, started three and four weeks after the Codex run, reach 1.059 and 1.063. Across the three Pi runs, the mean final score is 1.057 with a standard deviation of 0.007. Pi sessions cost less than half as much as Codex sessions on average ($1.02 versus $2.14) and are 15% shorter (9.0 versus 10.6 minutes).

**Ablating work session setups.** We compare three methods for managing agents' work sessions. In the default setting used throughout the paper, we launch short sessions consecutively and instruct agents to choose one task per session. We also consider alternatives with longer session, where each can last up to four hours, $25, or 1,000 turns, and agents can choose how many tasks to pursue. For these sessions, we compare two context-management methods: native compaction in Pi and Context Language Models (CLM; Shao et al., 2026), which let agents rewrite their working context. Figure 10(b) plots held-out speedup against API cost. At similar API costs, CLM performs similarly to the default setup (1.044 versus 1.049), and both score higher than native compaction (1.032).

## C.4 Budget-sharing implementation limitations

The instructions tell agents they can request shared sessions for themselves or another maintainer, but the budget system only accepts requests for the agent submitting them. Agents can still donate sessions from their own budgets to other maintainers. This mismatch prevents some approved requests from going through. For example, the system blocks a request for ten sessions for another maintainer in the default Opus run, even though three maintainers approve it.

Later requests or donations provide some of these sessions, so we cannot determine how much budget this restriction leaves unused by adding up the blocked requests. In the default Sol run, agents never request sessions for another maintainer; the final two requests fail because too few maintainers still have sessions left to vote. For Opus, the unused budget may reflect both the agents' decisions and this implementation restriction, and these runs do not establish how much each contributes.

# D Case Studies of Agent Behavior

## D.1 Communication experiment case studies

**A pool-key memo in `urllib3` (the opus-5 6 agent collaborative setup.** Every request through `urllib3` goes through one function that finds the connection pool for a host. In version 2.7.29 that function rebuilt the pool key from scratch on each call: it copied two dictionaries, lower-cased the scheme and host three times, and built a thirty-field tuple, only to look up a pool that was already open under the same key. None of `urllib3`'s own benchmarks run this function, so its maintainer had no way to see the cost by profiling its own code. The ISOLATED `urllib3` maintainer, same model and same starting code, never touched it.

The cost is visible from the caller's side. The `requests` maintainer timed its own request path with the socket removed and found this function was the largest remaining block it did not own, about 6 of 49 microseconds per request. It opened an issue on `urllib3` with the measurement, a table of what memoizing each candidate function would save, and the reason it had not cached the pool on its own side: skipping the lookup would stop refreshing the pool's recency and let a live pool be evicted under in-flight work.

Twenty-three minutes later the `urllib3` maintainer opened a pull request that memoizes the derived key while keeping the lookup, and named that reason as the design constraint. It spelled out when the memo is exact and when it must be skipped, for example for any context that carries headers or socket options, since mutating those changes the key without changing the snapshot. A second consumer, `requests-gssapi`, reported that its workload pays the cost twice per call, since it re-sends each request after an authentication challenge, and caught an error in the pull request's description before release. The fix shipped in `urllib3` 2.7.30 about an hour after the issue was opened. The `requests` maintainer then measured the release at 5 microseconds off every request, about 10%, with the same absolute saving on a real socket.

The changed function is on the call path of all four held-out packages. We do not attribute a change in the held-out score to it, since another `urllib3` release landed an hour later and the next sweeps moved by less than the run-to-run variation. What the case establishes is narrower: the cost could only be seen from a caller's workload, the fix was shaped by what the caller knew about its own side, and the ISOLATED maintainer never found it.

**Feedback catches a bug in a Gemini optimization (single agent vs 6 agent collaboration).** In the connected Gemini run, the `requests` maintainer speeds up `Response.content` by reading the response body in one call. The package's own 751 tests pass, but the change bypasses `urllib3`'s content-length check, allowing incomplete response bodies to pass without an error. The `responses` maintainer tests every `requests` release and reports the bug with a reproduction 15 minutes later. The `requests` maintainer releases a fix 32 minutes after the report, preserving the speedup. In the isolated run, the `requests` maintainer makes the same change a day later and publishes 15 further releases without restoring the check. No consumer in that run tests those releases, and the held-out correctness checks do not cover this case. The comparison illustrates how testing by another maintainer can catch an error that both the package's own tests and the held-out checks miss.

## D.2 Ownership experiment case studies

**Three agents write the same change.** The ten Opus maintainers began simultaneously and each proposed a division of work, but none of the first eight plans assigned `databricks-sql-connector`. Within twelve minutes, four maintainers opened pull requests on the same part of that package. Three rewrote the same twenty lines in essentially the same way, with two of the commits made one second apart, and the fourth repeated one of those changes in a neighboring file. Six maintainers commented on or reviewed the four pull requests before one was closed and two were cut back.

*Records:* Opus, ten maintainers without assigned ownership, `databricks-sql-connector` pull requests 1–4, board proposals 1–8.

**A package wrongly declared unusable.** Building `requests-gssapi` requires a system library, and the agents' environment starts with an empty list of system packages, so a plain install reports that the library does not exist. In the ten-maintainer run, one maintainer declared the package unbuildable and advised the others to stay away. Another counted four maintainers who had reached the same conclusion and asked that no one spend more time checking. Nine minutes after the first report, a third maintainer refreshed the package list, installed the library, and ran the tests successfully. The first maintainer withdrew its report, remarking that the agreement "measures shared method, not truth." The four- and six-maintainer runs repeated the same mistake and the same correction, and in the six-maintainer run the first pull request on the package came 44 hours later.

*Records:* Opus, ten maintainers without assigned ownership, `requests-gssapi` issue 1 and pull request 2, board proposal 1, with the same episode in the runs with four and six maintainers without assigned ownership.

**One maintainer works on the package that matters least.** With a single Opus maintainer responsible for all six packages, 194 of its 297 merged pull requests went to `nilearn`, which no other package depends on. Only 20% went to `requests` and `urllib3`, against 47–54% for the teams without assigned packages and 29% for the team with them. After its last change to those two packages, the maintainer spent two more days on `nilearn` while the held-out score stayed flat.

*Records:* Opus, one maintainer of all six packages, merged pull requests by repository, compared with the Opus runs with six assigned maintainers and with four, six, and ten maintainers without assigned ownership.

### D.3 Budget-sharing experiment case studies

**Agents usually approve requests under both sets of instructions.** In the guided runs, Sol maintainers cast 126 votes to approve requests and 16 to reject them, while Opus maintainers cast 27 to approve and five to reject. Maintainers who vote to reject requests ask for corrections to session transfers while supporting the proposed work. As in the main text, these counts include repeated votes and votes that maintainers later change or withdraw.

**Maintainers donate sessions to support work in other packages.** Under the default instructions, Opus maintainers make seven donations totaling 21 sessions, while Sol maintainers make none. Under the guided instructions, Sol maintainers make two donations totaling 24 sessions, and Opus maintainers make three totaling ten. These counts include only completed donations. In the guided Opus run, the `requests-gssapi` maintainer donates four sessions to `urllib3` after identifying a possible optimization there. Donations also occur in the guided Gemini run, where the `databricks-sql-connector` maintainer gives three sessions to `requests` and two to `urllib3` near the end of the run. The Gemini examples do not constitute a complete donation count.